\PassOptionsToPackage{dvipsnames}{xcolor}
\documentclass{article}

\usepackage{iclr2027_conference,times}
\usepackage[utf8]{inputenc}
\usepackage[T1]{fontenc}
\usepackage{hyperref}
\usepackage{url}
\usepackage{makecell}
\usepackage{booktabs}
\usepackage{rotating}
\usepackage{amsfonts}
\usepackage{amsmath}
\usepackage{caption}

\usepackage{nicefrac}
\usepackage{xcolor}
\usepackage{kotex}
\usepackage{adjustbox}
\usepackage{multirow}
\usepackage{amssymb}
\usepackage{longtable}
\usepackage{comment}
\usepackage[english]{babel}
\usepackage[autostyle]{csquotes}
\usepackage{hhline}
\usepackage{tabu}
\usepackage{tablefootnote}
\usepackage{graphicx}
\usepackage{wrapfig}
\usepackage{array}
\usepackage{capt-of}
\usepackage{relsize}
\usepackage{colortbl}
\usepackage{floatpag}
\floatpagestyle{plain}
\usepackage{float}
\usepackage{threeparttable}
\usepackage{enumitem}
\setlist[enumerate,itemize]{leftmargin=1.5em}
\usepackage{placeins}
\usepackage{bm}
\usepackage{tikz}
\usetikzlibrary{positioning,arrows.meta,calc,fit,backgrounds}
\usepackage{nicematrix}

\definecolor{LightYellow}{rgb}{1.0,0.98,0.85}
\definecolor{MainPurple}{HTML}{A451E4}
\newcommand{\cmark}{\textcolor{ForestGreen}{\checkmark}}
\newcommand{\xmark}{\textcolor{BrickRed}{\ensuremath{\times}}}

\title{FinVerse: Toward More Realistic Evaluation for Financial Time-Series Forecasting}

\author{
\parbox{\linewidth}{\centering
Jaehoon Lee\textsuperscript{1},
Jun Seo\textsuperscript{1},
Seunghan Lee\textsuperscript{1},
Junhyeok Kang\textsuperscript{1},
Sangjun Han\textsuperscript{1}, \\
Sungdong Yoo\textsuperscript{1},
Minjae Kim\textsuperscript{1},
Tae Yoon Lim\textsuperscript{1},
Dongwan Kang\textsuperscript{1},
Hwanil Choi\textsuperscript{1}, \\
Soonyoung Lee\textsuperscript{1},
Hao Ni\textsuperscript{2},
Srijan Sood\textsuperscript{3},
Stefan Zohren\textsuperscript{4},
Chanyeol Choi, \\
Yongjae Lee\textsuperscript{7},
Zhangyang Wang\textsuperscript{5},
Lukasz Szpruch\textsuperscript{6},
Benjamin Bowring\textsuperscript{2},
Wonbin Ahn\textsuperscript{1}
} \\[5ex]
\makebox[\linewidth][c]{%
\begin{tabular}{c}
\textsuperscript{1}LG AI Research \quad
\textsuperscript{2}University College London \quad
\textsuperscript{3}J.P. Morgan Chase \\[0.4ex]
\textsuperscript{4}University of Oxford \quad
\textsuperscript{5}University of Texas at Austin \quad
\textsuperscript{6}University of Edinburgh \\[0.4ex]
\textsuperscript{7}Ulsan National Institute of Science and Technology
\end{tabular}%
} \\[4ex]
\makebox[\linewidth][c]{\texttt{\{jaehoon.lee,wonbin.ahn\}@lgresearch.ai}}
}
\iclrfinalcopy

\begin{document}

\maketitle

\lhead{Preprint. Under review at ICLR 2027}

\begin{abstract}
Time-series foundation models (TSFMs) are increasingly evaluated on broad benchmarks that apply uniform conventional forecast evaluation metrics across heterogeneous time-series. However, in the financial domain, decision-relevant evaluation requires going beyond conventional metrics to assess forecasts in terms of directional accuracy, cross-sectional ranking, and portfolio outcomes. To this end, we introduce FinVerse, a financial time-series benchmark, collecting a broad collection of 116,897 series with 171.1M observations. FinVerse organizes financial data according to their economic semantics and evaluates target series across point-wise forecasting, cross-sectional ranking, and portfolio backtesting. Across 43 public TSFMs, model rankings on FinVerse show only moderate agreement with those on the existing benchmark, GIFT-Eval (r=0.43), suggesting that generic forecasting benchmarks do not fully capture the aspects of forecast quality that matter for financial decision-making. Furthermore, substantial variation in model performance across financial time-series categories and evaluation metrics provides a more informative basis for model selection, enabling practitioners to identify suitable models according to their target series and downstream objectives.

\end{abstract}

%=====================================================================
% 1. Introduction
%=====================================================================
\section{Introduction}
\label{sec:introduction}

The recent success of large language models has renewed interest in foundation models across a broad range of domains. This trend has naturally extended to time-series modeling, where a growing body of work aims to build general time-series foundation models (TSFMs) from many sources \citep{das2024decoder,liu2024timer,woo2024unified,goswami2024moment,liu2025timerxl,taga2025timepfn,ansari2025chronos2}. By pretraining on large and heterogeneous collections of time-series, these models seek to capture transferable temporal patterns that generalize across domains.

As time-series foundation models advance, developing benchmarks for evaluating them becomes increasingly important. A central question in designing such benchmarks is whether strong benchmark performance translates into strong performance on the downstream decisions that rely on these forecasts. In other words, if a decision maker selects the best-performing model on a benchmark, should they expect it to perform well for the decisions they ultimately care about? Existing time-series benchmarks such as GIFT-Eval \citep{aksu2024gifteval}, however, evaluate heterogeneous time-series using uniform conventional forecasting metrics, such as the Mean Absolute Scaled Error (MASE) and Continuous Ranked Probability Score (CRPS). Recent evidence suggests that strong performance on such benchmarks does not necessarily translate into strong performance in downstream financial decision-making \citep{salykaufmann2026deeplearningfinancialtime}.

This discrepancy may arise for the following two reasons. First, conventional forecast accuracy metrics do not fully characterize the aspects of a forecast that matter for financial decision-making. For example, in stock forecasting, anticipating the direction or relative ranking of future returns may be more relevant to investment decisions than minimizing MASE or CRPS. Second, what constitutes a useful forecast varies substantially across financial time series, making a common evaluation criterion inadequate for capturing their distinct economic roles. For inflation indicators such as the Consumer Price Index (CPI), for instance, it may be important to predict not only the direction but also whether inflation is accelerating or decelerating.

\begin{figure}[t]
\centering
\includegraphics[width=\linewidth]{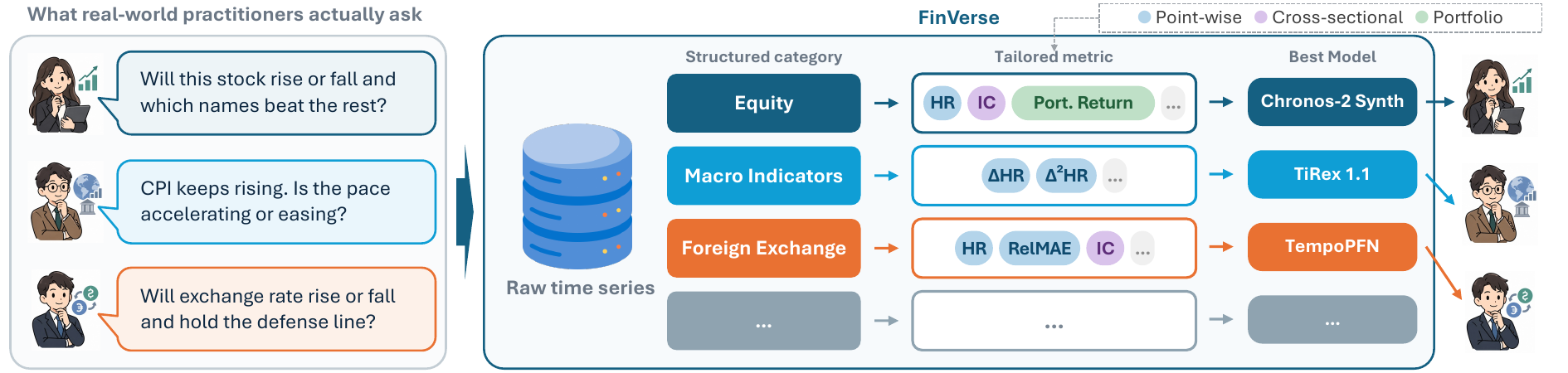}
\caption{Overview of FinVerse. FinVerse tailors evaluation metrics to each time series, providing a more informative basis for model selection based on the economic characteristics of the target series.}
\label{fig:main}
\end{figure}

% \begin{table}[t]
% \centering
% \caption{Coverage of the full FinVerse data artifact by scope and category. D, BD, W, M, Q, and A denote daily, business-daily, weekly, monthly, quarterly, and annual frequencies, respectively.}
% \label{tab:intro_data_summary}
% \vspace{0.35em}
% \small
% \setlength{\tabcolsep}{22pt}
% \renewcommand{\arraystretch}{1.08}
% \resizebox{\columnwidth}{!}{
% \begin{tabular}{llrrl}
% \toprule
% Scope & Category & Series & Points & Frequencies \\
% \midrule
% \multirow{6}{*}{Inter-country} & Foreign Exchange & 25 & 294.4K & BD \\
%  & Commodity & 59 & 261.9K & BD/W/M \\
%  & Crypto & 4,700 & 5.0M & D \\
%  & Global Liquidity & 39 & 4.5K & Q \\
%  & Global Risk & 6 & 18.1K & D/BD/M \\
%  & Market Index & 22 & 166.4K & BD \\
% \midrule
% \multirow{3}{*}{Country} & Fixed Income & 75 & 380.9K & BD/W/M \\
%  & Market Indicators & 32 & 270.8K & D/BD/W/M \\
%  & Macro Indicators & 199 & 165.2K & D/BD/W/M/Q/A \\
% \midrule
% \multirow{4}{*}{Individual} & ETF & 25,531 & 40.5M & BD \\
%  & Equity & 30,745 & 121.7M & BD \\
%  & Fundamentals & 55,442 & 2.2M & Q \\
%  & News Sentiment & 22 & 75.0K & D \\
% \bottomrule
% \end{tabular}
% }
% \end{table}

\begin{table}[t]
\centering
\caption{Coverage of the full FinVerse data by scope and category. D, BD, W, M, Q, and A denote daily, business-daily, weekly, monthly, quarterly, and annual frequencies, respectively. See Table~\ref{tab:appendix_detail_coverage} in the Appendix for detailed statistics at the detail-category level.}
\label{tab:intro_data_summary}
\small
\setlength{\tabcolsep}{30pt}
\resizebox{\columnwidth}{!}{
\begin{tabular}{llrl}
\toprule
Scope & Category & Points & Frequencies \\
\midrule
\multirow{6}{*}{Inter-country} & Foreign Exchange & 294.4K & BD \\
 & Commodity & 261.9K & BD/W/M \\
 & Crypto & 5.0M & D \\
 & Global Liquidity & 4.5K & Q \\
 & Global Risk & 18.1K & D/BD/M \\
 & Market Index & 166.4K & BD \\
\midrule
\multirow{3}{*}{Country} & Fixed Income & 380.9K & BD/W/M \\
 & Market Indicators & 270.8K & D/BD/W/M \\
 & Macro Indicators & 165.2K & D/BD/W/M/Q/A \\
\midrule
\multirow{4}{*}{Individual} & ETF & 40.5M & BD \\
 & Equity & 121.7M & BD \\
 & Fundamentals & 2.2M & Q \\
 & News Sentiment & 75.0K & D \\
\bottomrule
\end{tabular}
}
\end{table}

To address these limitations, we introduce \textbf{FinVerse} (\textbf{Fin}ancial time-series benchmark uni\textbf{Verse}), a benchmark for economically informed evaluation of financial time-series forecasting. As shown in Figure~\ref{fig:main}, FinVerse is built on the principle that each time-series should be evaluated according to the criteria that its economic meaning makes relevant, rather than applying a common set of conventional forecast accuracy metrics across heterogeneous financial time-series. FinVerse organizes financial time-series according to their economic semantics and evaluates each target series using criteria aligned with its role in financial decision-making. This design extends evaluation beyond conventional forecast accuracy to capture complementary aspects of forecast quality through point-wise forecasting, cross-sectional ranking, and portfolio backtesting.

FinVerse contains 116,897 financial time-series with 171.1M observations. The data are organized into a three-level hierarchy consisting of \textit{scope}, \textit{category}, and \textit{detail category}. At the highest level, time-series are classified into three scopes—\textit{inter-country}, \textit{country}, and \textit{individual}—according to the geographic and economic scope. Within each scope, the series are further grouped into categories based on the type of financial time-series they represent, and each category is subsequently divided into more granular detail categories. Table~\ref{tab:intro_data_summary} summarizes composition of FinVerse across categories. After establishing this hierarchy, we select 60,232 economically meaningful target series as the fixed evaluation set, comprising 13.6M observations collected from January 1, 2020 onward. FinVerse also provides data collected before January 1, 2020 for foundation model pre-training and fine-tuning. 

In our experiments, we find that models that perform well under generic time-series benchmarks do not necessarily perform well on FinVerse, highlighting the need for financial-tailored benchmarks. Moreover, our experiments reveal that the relative performance of forecasting models changes substantially across financial time-series categories and evaluation criteria. This result suggests that FinVerse can support more informed model selection by helping practitioners choose models that are appropriate for both the characteristics of the target financial time series and the intended forecasting objective. The main contributions are as follows:

\begin{itemize}[itemsep=1pt, topsep=2pt, parsep=0pt]
    \item We identify a key limitation of existing time-series benchmarks for financial forecasting: applying uniform conventional forecasting metrics across heterogeneous time-series may not adequately capture performance relevant to downstream financial decisions.
    \item We introduce FinVerse, a large-scale financial time-series benchmark comprising 116,897 series and 171.1M observations. FinVerse organizes financial series according to their economic semantics and evaluates them using criteria aligned with their roles in financial decision-making.
    \item We show that models performing well on generic time-series benchmarks do not necessarily perform well on FinVerse, providing empirical evidence for the need for financially tailored evaluation beyond conventional forecasting metrics.
    \item We demonstrate substantial variation in model performance across financial time-series categories and evaluation criteria. These results provide a more informative basis for model selection, helping practitioners identify models suited to their target time series and forecasting objectives.
\end{itemize}

%=====================================================================
% 2. Related Work
%=====================================================================
\section{Related Work}
\label{sec:related_work}

% \subsection{Time-Series Foundation Models}
% \label{subsec:ts_foundation_models}

\textbf{Time-Series Foundation Models (TSFMs).} The foundation model paradigm has expanded from language to time series. Traditional forecasting models are often trained for a specific dataset, frequency, or forecasting horizon, which limits their ability to transfer across heterogeneous temporal data. In contrast, recent TSFMs aim to pretrain on large and diverse collections of time series and then support zero-shot or few-shot forecasting. Representative forecasting-oriented models include TimesFM~\citep{das2024decoder}, Chronos~\citep{ansari2024chronos,ansari2025chronos2}, Moirai~\citep{woo2024unified}, Timer~\citep{liu2024timer,liu2025timerxl}, and TimePFN~\citep{taga2025timepfn}, which are pretrained on large collections of heterogeneous time series. The dominant architectural paradigm among recent foundation models is the decoder-only Transformer, which adapts autoregressive sequence modeling from language models to temporal prediction. Whether this generality transfers to finance is an open question: a recent study of foundation models on global equity returns reports that generic pretrained models underperform in zero-shot and fine-tuned settings, while models pretrained on financial data improve both forecast accuracy and economic utility~\citep{rahimikia2025revisiting}. 

% Hao Ni: My paper on TSFM in finance might be relevant (https://arxiv.org/abs/2511.18578), which shows that zero-shot or fine tuned TSFM performs very poor in financial data. In this work, we also pretrained TSFM from scratch on large-scale financial data, and the model codes/checkpoint can be found at https://github.com/DeepIntoStreams/TSFM_Finance . I am curious how our pretrained model works on your benchmarking dataset. Like what you did, we also evaluate a number of different TSFMs (about 17) with zero shot.
% Wonbin: Thanks Hao, I wasn't aware of this paper. Will add it here, it fits the motivation well. Running your checkpoint on FinVerse should be straightforward since we only need forecasts per series/origin. I'll ask the team to try it this week, but honestly it may not make the deadline. If not, we'll include it in the next version.

% \subsection{Time-Series Benchmarks}
% \label{subsec:ts_benchmarks}

\textbf{Time-Series Benchmarks.} Several benchmark efforts have shaped modern time-series forecasting evaluation. TSLib-style long-sequence forecasting protocols popularized standardized comparisons across canonical datasets \citep{zhou2021informer}. The Monash Time Series Forecasting Archive broadened the evaluation landscape by collecting diverse real-world forecasting datasets under a unified archive and metadata format \citep{godahewa2021monash}. More recently, LOTSA, introduced together with Moirai, scaled pretraining and evaluation toward a large collection of heterogeneous time-series datasets, supporting the development of universal forecasting models \citep{woo2024unified}. GIFT-Eval is a representative and widely used benchmark for general time-series forecasting model evaluation \citep{aksu2024gifteval}. It provides a standardized evaluation suite over heterogeneous datasets and has become a practical reference point for comparing recent time-series foundation models. 

Recent benchmark efforts have also begun to systematize financial time-series forecasting evaluation. FinBench and FinTSB provide dedicated benchmarks for financial time-series forecasting \citep{pederzoli2026finbench,hu2025fintsb}, but their evaluation scopes are primarily centered on stock data which limits diversity in data coverage. However, FinVerse collects financial time series across multiple categories, including equity prices, exchange rates, commodities, crypto assets, fixed-income indicators, market indicators, macroeconomic indicators, ETFs, and fundamentals. It further supports multiple frequencies and evaluation horizons, aiming to provide a broader and more practically useful evaluation setting for financial forecasting models.

\section{FinVerse}
  \label{sec:data}

FinVerse is built as a finance-domain benchmark universe. The benchmark first organizes diverse financial signals into a hierarchical taxonomy defined by Scope, Category, and Detail Category. For each Detail Category--Frequency combination, FinVerse assigns evaluation metrics that reflect the semantic meaning and economic role of the corresponding time series. The metric assignments for each Detail Category–Frequency combination are summarized in Tables~\ref{tab:appendix_metric_assignment_intercountry_foreign_exchange}–\ref{tab:appendix_metric_assignment_individual_fundamentals}.

\subsection{Benchmark Universe Construction}
\label{subsec:time_series_storage}

FinVerse organizes financial time-series along a scope--category--detail category hierarchy. The benchmark contains three scopes. The inter-country scope covers cross-country global series, including Foreign Exchange, Commodity, Crypto, Global Liquidity, Global Risk, and Market Index categories.
The country scope covers country-level economic and market conditions, including Fixed Income, Macro Indicators, and Market Indicators.
The individual scope covers individual tradable instruments, firm-level targets, and sector-level signals, including ETF, Equity, Fundamentals, and News Sentiment categories.
%
% Table~\ref{tab:intro_data_summary} summarizes the coverage of FinVerse at the category level, while Appendix Table~\ref{tab:appendix_detail_coverage} provides a more granular breakdown at the detail-category level. 
Table~\ref{tab:intro_data_summary} summarizes the category-level coverage, while Appendix Table~\ref{tab:appendix_detail_coverage} provides the detail-category breakdown.
In total, FinVerse contains 3 scopes, 13 categories, and 101 detail categories with 116,897 financial series and 171.1M observations. FinVerse includes daily, business-daily, weekly, monthly, quarterly, and annual frequencies. Note that the country and individual scopes are centered on the U.S. market, the world's largest financial market, whereas the inter-country scope includes cross-country and global financial series.

After organizing the hierarchy, we manually select 60,232 target series for evaluation based on their relevance to decision-making. These series contribute 13.6M observations over the evaluation period starting January 1, 2020. The resulting evaluation set spans 3 scopes, 10 categories, and 67 detail categories. The largest portion of it consists of business-daily observations from equity (8.0M) and ETF (4.5M) series, reflecting the large number of individual instruments represented in these categories, followed by quarterly company fundamentals (0.8M) and daily crypto series (140.5K). Although country- and inter-country-level series, which are primarily available at monthly, quarterly, and annual frequencies, contain fewer observations, they remain essential components of the benchmark because they capture economically important macroeconomic and market conditions.

% Country-level macro, fixed-income, and market-indicator groups contribute fewer observations

% The selected targets are dominated by dense traded-market series, with 8.0M observations from equity series, 4.5M from ETF series, 0.8M from fundamentals, and 140.0K from crypto (evaluation period, 2020 onward). Country-level macro, fixed-income, and market-indicator groups contribute fewer observations, but they expand the benchmark beyond asset-price forecasting by adding rates, spreads, inflation, labor, activity, volatility, and other economically distinct targets. 

% The frequency distribution shows the same structure. Business-daily series dominate the selected evaluation targets with 16.0M observations, while quarterly series contribute 1.2M observations and daily series contribute 163.7K observations. Monthly, weekly, and annual groups are smaller in raw observation count but are important for evaluating lower-frequency economic and accounting signals. This coverage motivates the metric design described below: dense traded-market groups support point-wise, ranking, and portfolio evaluation, whereas sparse or lower-frequency economic and accounting groups require metrics that respect their sampling frequency and economic interpretation.

\subsection{Benchmark Evaluation Metrics}
\label{subsec:benchmark_evaluation_metrics}

FinVerse uses metric families that correspond to three complementary evaluation views: \textit{point-wise forecasting}, \textit{cross-sectional ranking}, and \textit{portfolio performance}. Point-wise forecasting assesses individual predictions. In contrast, cross-sectional ranking evaluates whether a model correctly captures the relative ordering among a set of comparable assets forecast at the same point in time. Finally, portfolio performance translates model predictions into portfolio decisions under one fixed strategy and evaluates their realized performance through backtesting. The first two views are economic proxies: they measure properties of a forecast that financial decisions typically depend on, rather than the utility of a decision itself. Only the third view observes a realized outcome, and then only for the particular strategy defined below. In this section, we simply describe the metrics used for each evaluation view and their conceptual relevance to financial forecasting. Detailed mathematical definitions and implementation procedures are provided in Appendix~\ref{app:metric}.

\subsubsection{Point-wise Forecasting}

\paragraph{Hit Ratio (HR).} HR measures directional correctness. HR is motivated by the fact that many financial decisions depend more on predicting the direction of movement than on reconstructing the exact future value. For example, in asset-price forecasting, correctly predicting whether the price will rise or fall after a given horizon is often more important than minimizing the level error. We also propose \(\Delta\mathrm{HR}\) and \(\Delta^2\mathrm{HR}\), which
extend this directional criterion to first- and second-order changes, respectively. These variants are particularly useful for macroeconomic indicators such as CPI: \(\Delta\mathrm{HR}\)  evaluates whether the model correctly predicts whether the year-over-year value will increase or decrease, whereas \(\Delta^2\mathrm{HR}\) evaluates whether it correctly predicts whether the rate of change will accelerate or decelerate relative to the previous period. These variants require the predicted and observed series to match multiple directional conditions simultaneously, as defined in Appendix~\ref{app:metric}. They should therefore not be interpreted as simple binary directional-accuracy measures with a chance level of \(0.5\).

% HR variants capture a property that financial decisions frequently rely on; they do not measure the value of any decision that uses the forecast.

\paragraph{Relative Mean Absolute Error (RelMAE).}
FinVerse uses RelMAE, defined as the mean absolute error of the forecast divided by that of a naive forecast evaluated on the same cells. This differs from MASE, whose denominator is the in-sample seasonal-naive error computed from the history preceding the forecast. We adopt RelMAE to provide a direct comparison between model forecasts and simple naive forecasts under the same evaluation window, avoiding dependence on the length and seasonal structure of the available history. For completeness, we also report results using the classical MASE definition in Appendix~\ref{app:mase}. We include error-magnitude metrics because, although directional correctness is often the primary objective in financial forecasting, predictions with unrealistic magnitudes are of limited practical value. For example, in foreign exchange forecasting, where price movements are typically small, a model that consistently predicts excessively large changes may achieve reasonable directional accuracy while remaining unsuitable for real-world use.

Three variants are considered. First, $\mathrm{RelMAE}_{rel}$ evaluates the accuracy of predicting relative changes (e.g., returns) against a no-change relative baseline. Second, $\mathrm{RelMAE}_{abs}$ evaluates the accuracy of predicting raw values against a no-change raw-value baseline defined over the same comparison window. Finally, $\mathrm{RelMAE}_{ori}$ evaluates raw-value forecasts against the conventional origin-value persistence baseline, where the most recently observed value is repeated across all future horizons. While $\mathrm{RelMAE}_{rel}$ and $\mathrm{RelMAE}_{abs}$ are evaluated only at financially meaningful forecasting horizons (e.g., one week, one month, one quarter, half a year, and one year ahead), $\mathrm{RelMAE}_{ori}$ measures forecasting error over the entire prediction horizon. Results using alternative baselines beyond persistence are also provided in Appendix~\ref{app:alt_relmae}.

% Three variants are considered. First, \(\mathrm{RelMAE}_{rel}\) measures forecast error on relative changes against a no-change relative baseline. Second, \(\mathrm{RelMAE}_{abs}\) evaluates raw-value forecasts against a no-change raw-value baseline defined over the same comparison window. $\mathrm{RelMAE}_{rel}$ evaluates the accuracy of predicting relative changes (e.g., returns), whereas $\mathrm{RelMAE}_{abs}$ measures the accuracy of predicting the raw values themselves. Finally, \(\mathrm{RelMAE}_{ori}\) evaluates raw-value forecasts against the conventional origin-value persistence baseline, where the most recently observed value is repeated across all future horizons. Furthermore, $\mathrm{RelMAE}_{rel}$ and $\mathrm{RelMAE}_{abs}$ are evaluated only at financially meaningful forecasting horizons (e.g., one week, one month, one quarter, half a year, and one year ahead), while $\mathrm{RelMAE}_{ori}$ measures forecasting error over the entire prediction horizon. Results using alternative baselines beyond persistence are also provided in Appendix~\ref{app:additional_result_tables}.

 % Accordingly, point-wise evaluation should consider both directional behavior and magnitude accuracy.
% These three RelMAE variants capture complementary aspects of financial forecasting.  Furthermore, $\mathrm{RelMAE}_{rel}$ and $\mathrm{RelMAE}_{abs}$ are evaluated only at financially meaningful forecasting horizons (e.g., one week, one month, one quarter, half a year, and one year ahead), while $\mathrm{RelMAE}_{ori}$ measures forecasting error over the entire prediction horizon.

\subsubsection{Cross-sectional Ranking}

The information coefficient evaluates cross-sectional ranking quality. For a set of assets, FinVerse computes a Spearman rank correlation between predicted and realized future changes. IC is selected because portfolio and allocation decisions often depend on relative ordering across assets rather than independent point accuracy for each series. Therefore, it serves as a proxy for that ordering quality rather than as a measure of allocation outcomes. This is used for groups where relative ranking is meaningful, such as currencies, commodities, crypto assets, global market indices, ETFs, equities, selected fixed-income groups, and selected fundamentals.

\subsubsection{Portfolio Performance}
FinVerse also evaluates how forecasts behave once they are translated into portfolio decisions, converting forecasts into an equal-weight long-only strategy. At each rebalancing point, assets are ranked by their predicted signal, and the top 10\% are selected to construct an equal-weight portfolio, with a minimum of ten assets. The selected portfolio is then held for the  corresponding horizon before being rebalanced. We consider holding horizons of one week, one month, one quarter, half a year, and one year, restricted in each group to the horizons its sampling frequency admits. To reduce sensitivity to the choice of a calendar start date, FinVerse performs the backtest using every possible starting offset within each holding horizon and averages the resulting portfolio statistics.

% FinVerse also evaluates how forecasts behave once they are translated into portfolio decisions, converting forecasts into a simple equal-weight long-only strategy. At each rebalancing point, assets are ranked by their predicted signal, and the top 10\% are selected to construct an equal-weight portfolio, with a minimum of ten assets. The selected portfolio is then held for the corresponding horizon before being rebalanced. We consider holding horizons of one week, one month, one quarter, half a year, and one year, restricted in each group to the horizons its sampling frequency admits. To reduce sensitivity to the choice of a particular calendar start date, FinVerse performs the backtest using every possible starting offset within each holding horizon and averages the resulting portfolio statistics. 

% For example, for a one-month holding horizon, portfolios starting on different days are evaluated as separate monthly rebalancing paths rather than relying on a single arbitrarily chosen start date. 

For each portfolio, we measure annualized return, annualized volatility, Sharpe ratio, and maximum drawdown (MDD). These metrics capture complementary aspects of the realized investment outcomes. Annualized return measures portfolio profitability, while annualized volatility captures the variability of realized returns. The Sharpe ratio evaluates return relative to risk, and MDD measures the largest peak-to-trough portfolio loss. The portfolio aspect rewards lower risk as well as higher returns. We acknowledge that portfolio performance can depend on the choice of investment strategy. To start from a simple setting, we keep the portfolio construction assumptions minimal, following a setup similar to \citet{pederzoli2026finbench}. Details are described in Appendix~\ref{app:assume_port}. 

% We leave richer strategy families for future extensions.

% ; FinVerse therefore adopts a simple and transparent portfolio construction rule as a common evaluation setting, while leaving richer strategy families for future extensions.

\subsection{Evaluation Periods and Forecast Horizons}
\label{subsec:evaluation_periods_horizons}

\begin{wrapfigure}{r}{0.41\linewidth}
\vspace{-1.0em}
\centering
\captionsetup{type=table}
\caption{Prediction lengths and minimum lookback requirements per frequency.}
\label{tab:freq_horizon_lookback}

\small
\setlength{\tabcolsep}{1.5pt}
\renewcommand{\arraystretch}{1.05}

\begin{tabular}{@{}lcc@{}}
\toprule
Frequency & Pred. Length & Min. Lookback \\
\midrule
Daily          & 365 & 365 \\
Business-daily & 260 & 260 \\
Weekly         & 52  & 52  \\
Monthly        & 12  & 12  \\
Quarterly      & 4   & 4   \\
Annual         & 1   & 4   \\
\bottomrule
\end{tabular}
\vspace{-0.5em}
\end{wrapfigure}

We construct the evaluation test set using observations from the evaluation start date, \texttt{2020-01-01}, onward. As summarized in Table~\ref{tab:freq_horizon_lookback}, we define the prediction length and the minimum lookback window size for each sampling frequency. The prediction lengths are chosen to correspond to financially meaningful forecasting horizons (i.e., one week, one month, one quarter, half a year, and one year ahead), while the minimum lookback window sizes are selected to provide sufficient historical context for making meaningful forecasts at the corresponding horizons.

\begin{figure}[t]
\centering
\includegraphics[width=\linewidth]{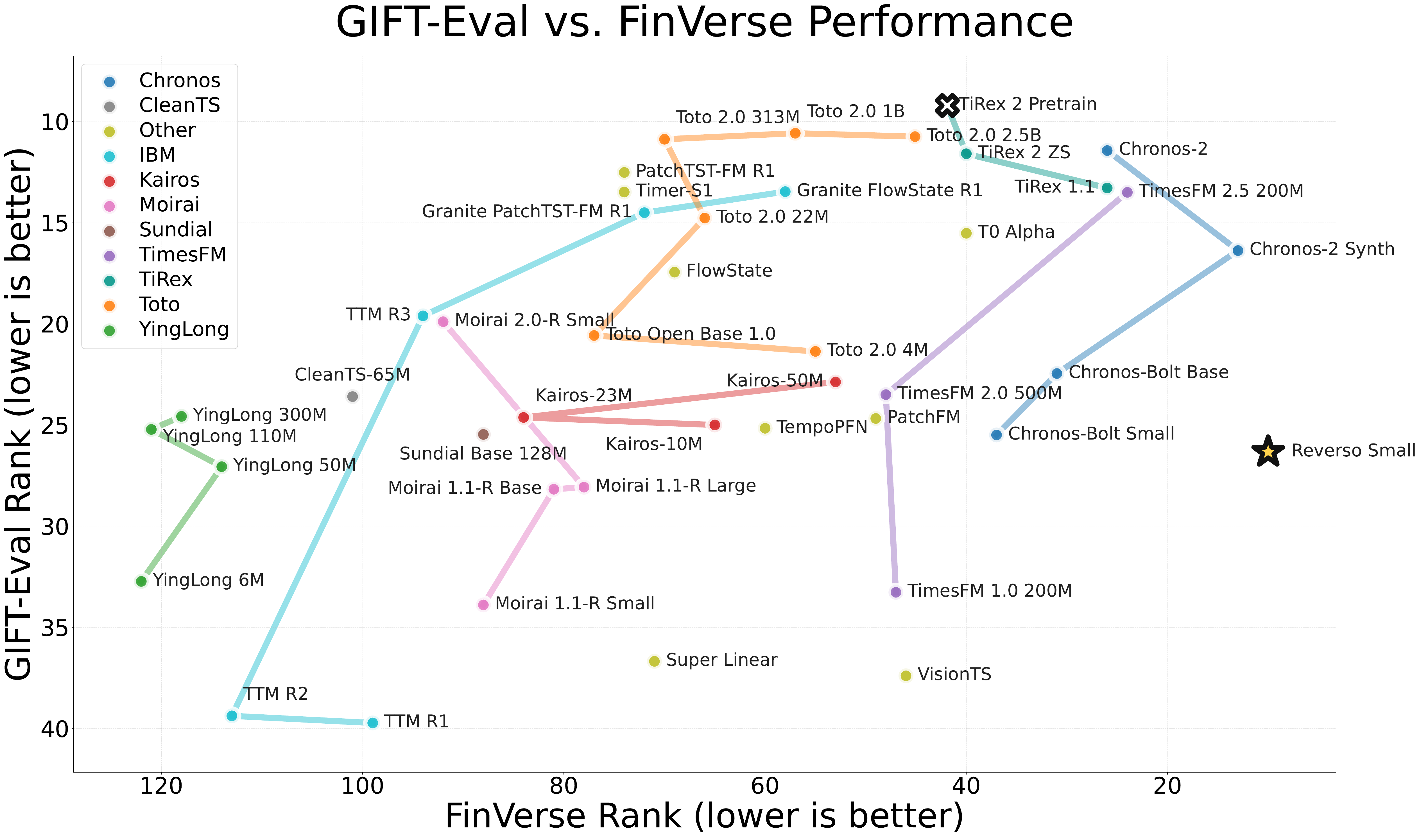}
\caption{Comparison between FinVerse and GIFT-Eval rankings. The GIFT-Eval rank is the average of the public GIFT-Eval MASE rank and probabilistic forecast rank. The star marks the best-ranked model in FinVerse, and the X marker marks the best-ranked model in GIFT-Eval.}
\label{fig:gifteval_finverse_rank}
\end{figure}

\section{Experiments}
\label{sec:evaluation_with_timeverse_fin}

\subsection{Experimental Setup}
% \label{subsec:foundation_models}

% Hao Ni: so all the models are zero-shot without any fine tuning? 
% Wonbin: Yes, all zero-shot with the public checkpoints, no fine-tuning from our side. (TiRex 2 Pretrain vs ZS are just two different released checkpoints.) We release a pre-2020 training pool but didn't use it for any of the reported numbers. Good point that this isn't said anywhere, will add a sentence.

\textbf{Foundation Models.} We evaluate FinVerse with 43 time-series foundation models, covering major model families such as TimesFM~\citep{das2024decoder}, Chronos~\citep{ansari2024chronos,ansari2025chronos2}, Moirai~\citep{woo2024unified}, Timer~\citep{liu2024timer,liu2025timerxl}, Toto~\citep{toto2025}, TiRex~\citep{dooley2025tirex}, and TTM~\citep{ekambaram2024tinytimemixer}. The complete list of evaluated models is provided in Appendix~\ref{app:foundation_list}. All foundation models are evaluated in a zero-shot setting without any fine-tuning on FinVerse. However, given the release dates of the evaluated models and the fact that the FinVerse evaluation period begins in 2020, we cannot completely rule out the possibility that some evaluation-period data may have been included in their pretraining data. For all models, we use the minimum lookback window specified for each frequency in Table~\ref{tab:freq_horizon_lookback} as a context window.

% We evaluate FinVerse with 43개 time-series foundation models. 대표 적인 군집... chronose, timesfm 등등 너가 적어주고. Refer to Appendix~\ref{app:foundation_list} to detail. All foundation models are evaluated in a zero-shot setting without any fine-tuning on FinVerse. However, given the release dates of the evaluated models and the fact that the FinVerse evaluation period begins in 2020, we cannot completely rule out the possibility that some evaluation-period data may have been included in their pretraining data. For all models, we use the minimum lookback window specified for each frequency in Table~\ref{tab:freq_horizon_lookback}.

% a broad set of time-series foundation models. 

% All foundation models are evaluated in a zero-shot setting without any fine-tuning on FinVerse. However, given the release dates of the evaluated models and the fact that the FinVerse evaluation period begins in 2020, we cannot completely rule out the possibility that some evaluation-period data may have been included in their pretraining data. For all models, we use the minimum lookback window specified for each frequency in Table~\ref{tab:freq_horizon_lookback}.

% For all models, we use the minimum lookback window specified for each frequency in Table~\ref{tab}. Because FinVerse evaluates time series from 2020 onward, the pretraining periods of some models may overlap with the evaluation period

\textbf{Overall Rank Aggregation.} FinVerse employs a diverse set of evaluation metrics, whose numerical magnitudes can vary substantially across financial time series. Directly averaging raw metric values would therefore overemphasize certain metrics and categories. To address this issue, FinVerse adopts a rank-based aggregation strategy rather than directly averaging raw metric values. Specifically, we first compute the average score of each evaluation metric within every detail category--frequency combination. Models are then ranked separately for each metric within each detail category--frequency combination. Next, for each of the three evaluation groups (i.e., Point-wise Forecasting, Cross-sectional Ranking, and Portfolio Performance), we compute the geometric mean of the corresponding metric ranks, denoted by \(K\). The models are then ranked according to the \(K\) value of each evaluation group. Finally, the overall rank is obtained by summing the three group-level ranks. This aggregation strategy enables heterogeneous metrics to contribute evenly to the final evaluation while preventing large categories from dominating the benchmark. As a result, smaller but economically important categories are appropriately reflected in the overall ranking.

\subsection{Overall Benchmark Performance}

Figure~\ref{fig:gifteval_finverse_rank} compares the overall FinVerse rank with the corresponding GIFT-Eval ranking for models evaluated on both benchmarks. Across the 43 matched models, the two rankings agree only moderately (Pearson \(r=0.43\); Spearman \(\rho=0.41\); Kendall \(\tau=0.27\)), indicating that they are related but far from interchangeable. 
% Because the comparison is between rankings, we report the rank correlations alongside the linear one; all three agree that the association is weak. 
For example, Reverso Small is the strongest FinVerse model despite a weaker GIFT-Eval rank. The result shows that strong performance on a generic benchmark does not necessarily translate into superior performance under our decision-relevant financial evaluation.

% 's decision-relevant financial evaluation.

\subsection{Performance Across Evaluation Groups}

\begin{figure}[t]
\centering
\includegraphics[width=\linewidth]{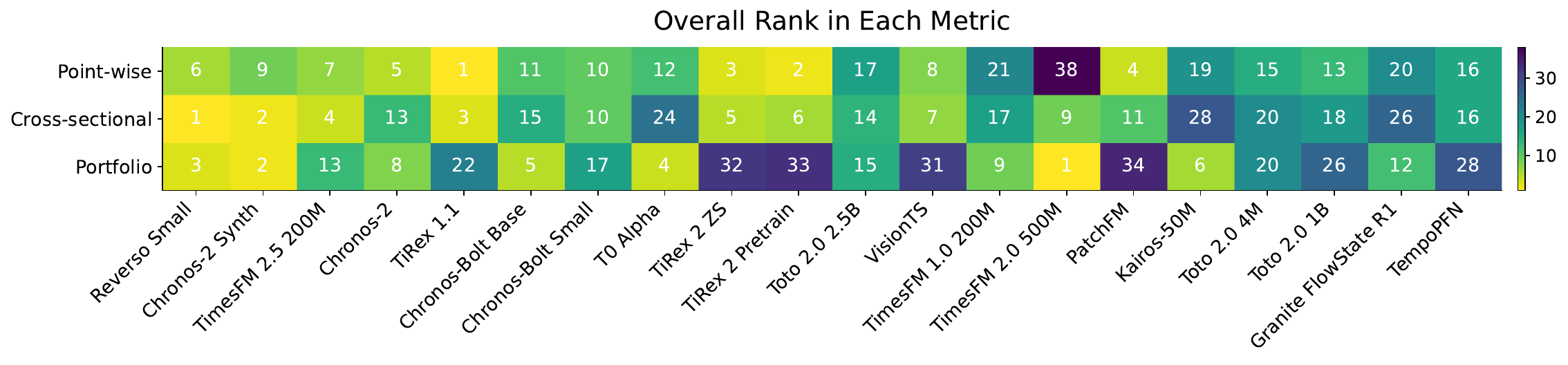}
\caption{Top-20 models ranked separately under three aspects: point-wise forecasting, cross-sectional ranking, and portfolio performance aspects.}
\label{fig:aspect_rank_heatmap}
\end{figure}

\begin{figure}[t]
\centering
\includegraphics[width=\linewidth]{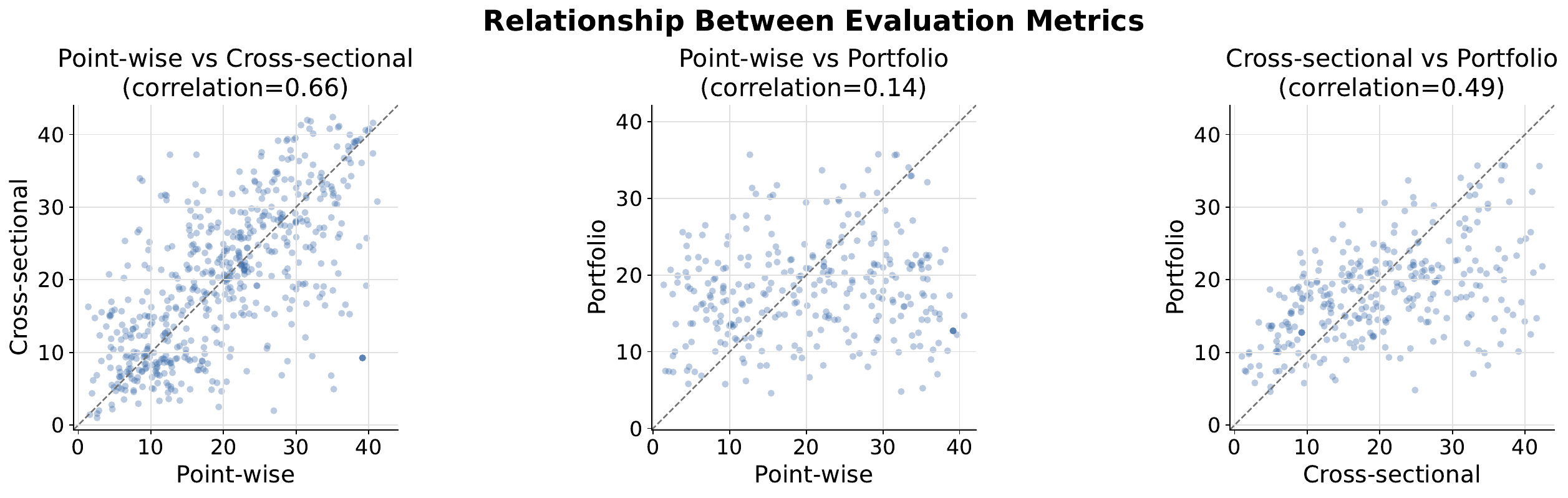}
\caption{Pairwise relationship between point-wise, cross-sectional, and portfolio ranks.}
\label{fig:metric_relationship_scatter}
\end{figure}

Figure~\ref{fig:aspect_rank_heatmap} reports model rankings separately for the three evaluation aspects: point-wise forecasting, cross-sectional ranking, and portfolio performance. TiRex 1.1 performs best in the point-wise aspect, followed by TiRex 2 Pretrain. Cross-sectional ranking favors Reverso Small and Chronos-2 Synth. Portfolio performance also favors different models: TimesFM 2.0 500M ranks first, followed by Chronos-2 Synth and Reverso Small. 
% These results show that model rankings vary across evaluation aspects. 
Figure~\ref{fig:metric_relationship_scatter} further examines the relationships among the three evaluation aspects at the detail-category level, where each point is one model evaluated on one detail category--frequency unit and its coordinate is the geometric mean of its ranks under that aspect. Point-wise and cross-sectional ranking are the most closely related pair (\(r=0.66\); \(\rho=0.67\); \(\tau=0.49\)). The point-wise--portfolio relationship is much weaker (\(r=0.14\), \(\rho=0.12\), \(\tau=0.08\)), while the cross-sectional--portfolio relationship is moderately aligned (\(r=0.49\), \(\rho=0.48\), \(\tau=0.35\)). Portfolio results therefore carry information that conventional per-series accuracy does not, and the three aspects are complementary rather than redundant, supporting a benchmark that evaluates forecasting models from three perspectives.

% These results demonstrate that strong point-wise forecasting performance does not necessarily imply superior cross-sectional ranking or portfolio performance.

\begin{table}[t]
\setlength{\tabcolsep}{2pt}
\centering
\caption{Category-wise rank aggregation. Lower values are better.}
\label{tab:categorywise_model_performance}
\resizebox{\linewidth}{!}{
\begin{tabular}{@{}lcccccccccc@{}}
\toprule
\multirow{3}{*}{Model} & \multicolumn{4}{c}{Inter-country} & \multicolumn{3}{c}{Country} & \multicolumn{3}{c}{Individual} \\
\cmidrule(lr){2-5}\cmidrule(lr){6-8}\cmidrule(lr){9-11}
 & Commodity & Crypto & \makecell{Foreign\\Exchange} & \makecell{Market\\Index} & \makecell{Fixed\\Income} & \makecell{Macro\\Indicators} & \makecell{Market\\Indicators} & ETF & Equity & Fundamentals \\
\midrule
Reverso Small & 8.80 & 11.50 & 17.50 & 6.37 & \textbf{8.33} & 9.47 & 11.26 & 9.71 & 6.12 & 10.55 \\
Chronos-2 Synth & \textbf{6.78} & \textbf{4.34} & 8.08 & 10.10 & 11.08 & 12.54 & 13.47 & 6.34 & \textbf{4.07} & 13.85 \\
TimesFM 2.5 200M & 11.90 & 12.80 & 7.43 & 11.36 & 11.54 & 9.58 & 10.48 & 11.66 & 11.25 & 10.53 \\
Chronos-2 & 11.46 & 11.55 & 13.75 & 12.70 & 13.98 & 8.41 & 11.29 & 10.11 & 13.46 & 9.35 \\
TiRex 1.1 & 10.47 & 14.31 & 12.30 & 18.06 & 9.22 & \textbf{5.50} & 10.70 & 11.95 & 12.08 & 8.29 \\
Chronos-Bolt Base & 14.76 & 11.41 & 10.26 & 13.39 & 11.97 & 12.03 & 16.13 & 10.39 & 16.41 & 11.69 \\
Chronos-Bolt Small & 14.70 & 15.09 & 10.70 & 10.52 & 12.20 & 12.03 & 14.78 & 10.91 & 18.39 & 11.55 \\
T0 Alpha & 19.68 & 24.04 & 14.45 & 20.85 & 12.45 & 11.00 & 23.36 & 17.58 & 11.24 & 15.72 \\
TiRex 2 ZS & 12.26 & 13.09 & 15.61 & 19.43 & 10.41 & 6.85 & 15.95 & 20.66 & 22.89 & 8.13 \\
TiRex 2 Pretrain & 11.50 & 10.06 & 13.30 & 17.41 & 9.98 & 5.96 & 12.44 & 19.61 & 21.54 & 9.06 \\
Toto 2.0 2.5B & 7.31 & 13.50 & 10.37 & 10.19 & 9.56 & 16.82 & \textbf{2.92} & \textbf{3.35} & 8.62 & 21.73 \\
VisionTS & 22.91 & 23.60 & 9.95 & 16.30 & 12.27 & 10.69 & 26.10 & 20.95 & 24.57 & \textbf{6.73} \\
TimesFM 1.0 200M & 20.34 & 17.23 & 13.88 & 24.36 & 19.91 & 15.99 & 13.26 & 16.74 & 18.86 & 13.30 \\
TimesFM 2.0 500M & 21.88 & 14.80 & 16.07 & 27.83 & 29.84 & 30.33 & 32.91 & 12.37 & 11.95 & 15.11 \\
PatchFM & 12.34 & 14.56 & 17.38 & 11.26 & 8.51 & 7.40 & 11.15 & 13.75 & 11.33 & 19.99 \\
Kairos-50M & 12.97 & 20.87 & 23.92 & 11.33 & 14.25 & 15.19 & 11.13 & 8.69 & 16.89 & 22.72 \\
Toto 2.0 4M & 10.71 & 12.78 & 18.46 & 12.52 & 13.31 & 12.97 & 13.23 & 13.27 & 13.75 & 21.90 \\
Toto 2.0 1B & 6.93 & 12.35 & 11.49 & 11.98 & 10.62 & 15.24 & 5.45 & 5.04 & 7.37 & 21.52 \\
Granite FlowState R1 & 12.82 & 22.54 & 20.15 & 15.47 & 13.71 & 15.13 & 20.68 & 19.57 & 21.18 & 23.44 \\
TempoPFN & 9.48 & 21.09 & \textbf{4.39} & \textbf{4.66} & 10.63 & 16.27 & 4.24 & 8.40 & 11.32 & 21.07 \\
Kairos-10M & 15.60 & 21.05 & 17.30 & 10.95 & 15.17 & 20.73 & 19.47 & 15.53 & 18.71 & 29.76 \\
Toto 2.0 22M & 11.29 & 9.63 & 15.60 & 11.34 & 14.05 & 15.57 & 7.50 & 13.05 & 15.13 & 19.55 \\
FlowState & 13.05 & 24.62 & 21.26 & 15.41 & 15.26 & 15.29 & 20.52 & 20.25 & 18.22 & 27.98 \\
Toto 2.0 313M & 8.42 & 11.59 & 14.28 & 12.51 & 12.80 & 15.03 & 5.62 & 8.22 & 10.12 & 19.15 \\
Super Linear & 22.02 & 21.74 & 14.39 & 22.31 & 16.43 & 18.99 & 18.09 & 27.86 & 21.11 & 9.86 \\
Granite PatchTST-FM R1 & 27.73 & 20.26 & 22.26 & 24.09 & 31.80 & 29.56 & 29.03 & 28.57 & 20.63 & 19.57 \\
PatchTST-FM R1 & 27.92 & 21.58 & 21.77 & 22.51 & 25.60 & 26.30 & 28.09 & 27.87 & 19.46 & 19.37 \\
Timer-S1 & 20.64 & 19.48 & 19.82 & 26.10 & 18.87 & 18.51 & 23.76 & 28.82 & 27.65 & 12.24 \\
Toto Open Base 1.0 & 15.94 & 15.74 & 34.67 & 12.78 & 14.97 & 20.40 & 22.85 & 17.68 & 17.74 & 16.77 \\
Moirai 1.1-R Large & 28.64 & 13.74 & 30.07 & 25.02 & 22.37 & 28.54 & 15.25 & 16.28 & 12.83 & 23.67 \\
Moirai 1.1-R Base & 27.51 & 18.48 & 24.52 & 28.04 & 20.11 & 27.86 & 26.41 & 20.10 & 18.98 & 21.15 \\
Kairos-23M & 22.03 & 23.89 & 14.87 & 22.44 & 23.31 & 19.38 & 23.53 & 18.05 & 23.07 & 24.41 \\
Moirai 1.1-R Small & 27.48 & 18.67 & 19.62 & 26.56 & 21.55 & 25.57 & 22.56 & 26.69 & 23.51 & 19.88 \\
Sundial Base 128M & 23.14 & 18.19 & 22.04 & 27.22 & 20.95 & 18.86 & 26.08 & 31.70 & 27.81 & 14.91 \\
Moirai 2.0-R Small & 25.99 & 20.48 & 25.88 & 21.52 & 23.27 & 21.78 & 25.70 & 23.10 & 17.61 & 22.68 \\
TTM R3 & 26.50 & 25.65 & 27.38 & 16.74 & 27.53 & 28.44 & 26.77 & 26.87 & 28.10 & 17.36 \\
TTM R1 & 24.00 & 24.81 & 21.84 & 29.99 & 27.58 & 22.91 & 25.30 & 30.03 & 23.43 & 16.58 \\
CleanTS-65M & 28.08 & 22.42 & 24.91 & 26.17 & 30.52 & 25.76 & 24.09 & 34.51 & 25.36 & 22.51 \\
TTM R2 & 27.51 & 21.47 & 24.62 & 15.61 & 23.56 & 21.41 & 29.67 & 29.53 & 28.49 & 15.76 \\
YingLong 50M & 28.97 & 22.21 & 21.05 & 26.70 & 35.04 & 36.58 & 33.95 & 33.40 & 27.99 & 26.73 \\
YingLong 300M & 29.44 & 22.01 & 24.44 & 26.71 & 35.70 & 36.70 & 33.01 & 33.19 & 26.16 & 26.88 \\
YingLong 110M & 28.92 & 22.42 & 25.70 & 26.91 & 35.40 & 36.56 & 32.65 & 33.28 & 27.22 & 27.50 \\
YingLong 6M & 29.77 & 23.74 & 24.76 & 25.27 & 33.39 & 36.59 & 31.82 & 33.85 & 26.83 & 26.84 \\
\bottomrule
\end{tabular}
}
\end{table}

\begin{table}[t]
\setlength{\tabcolsep}{1pt}
\centering
% \caption{Metric-wise model performance across point-wise, cross-sectional, and portfolio-oriented metrics using averaged raw metric values.}
\caption{Metric-wise model performance across point-wise, cross-sectional, and portfolio-oriented metrics using averaged raw metric values. When we calculate the Sharpe ratio, we do not subtract the risk-free rate and instead use raw portfolio returns.
Refer to Appendix~\ref{app:reason_sharpe} for details on why the reported Sharpe ratio does not equal the ratio of the aggregated return and volatility values.}
\label{tab:metricwise_model_performance}
\resizebox{\linewidth}{!}{
\begin{tabular}{@{}lccccccccccc@{}}
\toprule
Model & HR$\uparrow$ & $\Delta$HR$\uparrow$ & $\Delta^2$HR$\uparrow$ &
$\mathrm{RelMAE}_{rel}\downarrow$ & $\mathrm{RelMAE}_{abs}\downarrow$ &
$\mathrm{RelMAE}_{ori}\downarrow$ & IC$\uparrow$ &
Return$\uparrow$ & Sharpe$\uparrow$ & MDD$\uparrow$ & Vol.$\downarrow$ \\
% Model & HR & $\Delta$HR & $\Delta^2$HR & $\mathrm{RelMAE}_{rel}$ & $\mathrm{RelMAE}_{abs}$ & $\mathrm{RelMAE}_{ori}$ & IC & Return & Sharpe & MDD & Vol. \\
% Model & HR$(\uparrow)$ & $\Delta$HR$(\uparrow)$ & $\Delta^2$HR$(\uparrow)$ &
% $\mathrm{RelMAE}_{rel}(\downarrow)$ & $\mathrm{RelMAE}_{abs}(\downarrow)$ &
% $\mathrm{RelMAE}_{ori}(\downarrow)$ & IC$(\uparrow)$ &
% Return$(\uparrow)$ & Sharpe$(\uparrow)$ & MDD$(\downarrow)$ & Vol.$(\downarrow)$ \\
\midrule
Reverso Small & \textbf{0.664} & 0.545 & 0.484 & 0.851 & 0.849 & \textbf{0.981} & 0.123 & 0.171 & 0.697 & -0.349 & 0.470 \\
Chronos-2 Synth & 0.647 & 0.557 & 0.500 & 0.879 & 0.867 & 1.020 & \textbf{0.147} & 0.254 & 0.701 & -0.349 & 0.705 \\
TimesFM 2.5 200M & 0.648 & 0.568 & 0.512 & 0.855 & 0.849 & 1.002 & 0.130 & 0.155 & 0.686 & -0.373 & 0.478 \\
Chronos-2 & 0.652 & 0.562 & 0.499 & 0.843 & 0.848 & 1.012 & 0.054 & 0.208 & 0.660 & -0.365 & 0.650 \\
TiRex 1.1 & 0.663 & \textbf{0.576} & \textbf{0.520} & \textbf{0.835} & \textbf{0.833} & 0.992 & 0.135 & 0.174 & 0.674 & -0.369 & 0.596 \\
Chronos-Bolt Base & 0.638 & 0.560 & 0.494 & 0.879 & 0.878 & 1.049 & 0.073 & 0.165 & 0.706 & -0.365 & 0.426 \\
Chronos-Bolt Small & 0.638 & 0.554 & 0.493 & 0.886 & 0.882 & 1.063 & 0.072 & 0.210 & 0.680 & -0.382 & 0.623 \\
T0 Alpha & 0.624 & 0.571 & 0.511 & 0.934 & 0.930 & 1.207 & 0.119 & 0.251 & 0.669 & -0.365 & 0.766 \\
TiRex 2 ZS & 0.649 & 0.571 & 0.517 & 0.838 & 0.837 & 1.007 & 0.133 & 0.166 & 0.640 & -0.382 & 0.591 \\
TiRex 2 Pretrain & 0.651 & 0.573 & 0.519 & 0.836 & \textbf{0.833} & 1.001 & 0.134 & 0.180 & 0.631 & -0.383 & 0.625 \\
Toto 2.0 2.5B & 0.651 & 0.546 & 0.488 & 0.961 & 0.936 & 1.139 & 0.098 & 0.205 & 0.679 & -0.372 & 0.665 \\
VisionTS & 0.634 & 0.567 & 0.506 & 0.900 & 0.907 & 1.066 & 0.087 & 0.214 & 0.680 & -0.401 & 0.669 \\
TimesFM 1.0 200M & 0.624 & 0.559 & 0.502 & 0.947 & 0.942 & 1.159 & 0.120 & 0.227 & 0.679 & -0.373 & 0.637 \\
TimesFM 2.0 500M & 0.536 & 0.431 & 0.376 & 1.347 & 1.303 & 1.341 & 0.113 & 0.140 & 0.665 & \textbf{-0.341} & 0.424 \\
PatchFM & \textbf{0.664} & 0.565 & 0.503 & 0.871 & 0.859 & 1.022 & 0.116 & 0.168 & 0.646 & -0.372 & 0.518 \\
Kairos-50M & 0.649 & 0.547 & 0.481 & 0.972 & 0.956 & 1.109 & 0.088 & 0.203 & 0.705 & -0.358 & 0.650 \\
Toto 2.0 4M & 0.652 & 0.555 & 0.493 & 0.919 & 0.912 & 1.084 & 0.097 & 0.215 & 0.678 & -0.367 & 0.678 \\
Toto 2.0 1B & 0.651 & 0.552 & 0.493 & 0.963 & 0.936 & 1.140 & 0.098 & 0.191 & 0.670 & -0.372 & 0.566 \\
Granite FlowState R1 & 0.634 & 0.565 & 0.490 & 1.022 & 0.990 & 1.240 & 0.042 & 0.214 & 0.702 & -0.382 & 0.599 \\
TempoPFN & 0.657 & 0.513 & 0.443 & 0.978 & 0.943 & 1.066 & 0.056 & 0.198 & 0.642 & -0.374 & 0.706 \\
Kairos-10M & 0.644 & 0.503 & 0.439 & 1.172 & 1.127 & 1.320 & 0.070 & 0.231 & \textbf{0.708} & -0.374 & 0.710 \\
Toto 2.0 22M & 0.650 & 0.553 & 0.490 & 0.954 & 0.935 & 1.131 & 0.099 & 0.213 & 0.655 & -0.367 & 0.564 \\
FlowState & 0.632 & 0.553 & 0.480 & 1.012 & 0.981 & 1.207 & 0.024 & 0.244 & 0.687 & -0.379 & 0.739 \\
Toto 2.0 313M & 0.652 & 0.556 & 0.492 & 0.952 & 0.933 & 1.120 & 0.091 & 0.189 & 0.671 & -0.370 & 0.596 \\
Super Linear & 0.599 & 0.548 & 0.486 & 0.952 & 0.941 & 1.116 & 0.109 & 0.255 & 0.670 & -0.401 & 0.773 \\
Granite PatchTST-FM R1 & 0.536 & 0.440 & 0.381 & 1.423 & 1.400 & 1.473 & 0.096 & 0.245 & 0.635 & -0.385 & 0.732 \\
PatchTST-FM R1 & 0.564 & 0.448 & 0.388 & 1.303 & 1.260 & 1.334 & 0.091 & 0.249 & 0.634 & -0.387 & 0.755 \\
Timer-S1 & 0.610 & 0.541 & 0.482 & 0.968 & 0.953 & 1.166 & 0.120 & 0.149 & 0.634 & -0.397 & 0.608 \\
Toto Open Base 1.0 & 0.596 & 0.504 & 0.442 & 0.995 & 0.975 & 1.117 & 0.079 & 0.232 & 0.638 & -0.377 & 0.730 \\
Moirai 1.1-R Large & 0.531 & 0.447 & 0.394 & 1.247 & 1.180 & 1.406 & 0.096 & 0.160 & 0.626 & -0.356 & \textbf{0.373} \\
Moirai 1.1-R Base & 0.536 & 0.450 & 0.393 & 1.207 & 1.166 & 1.348 & 0.087 & 0.144 & 0.624 & -0.360 & 0.465 \\
Kairos-23M & 0.636 & 0.521 & 0.456 & 1.071 & 1.065 & 1.272 & 0.056 & 0.230 & 0.650 & -0.374 & 0.749 \\
Moirai 1.1-R Small & 0.544 & 0.454 & 0.401 & 1.191 & 1.145 & 1.275 & 0.105 & 0.156 & 0.652 & -0.380 & 0.461 \\
Sundial Base 128M & 0.607 & 0.546 & 0.487 & 0.986 & 0.972 & 1.200 & 0.110 & 0.225 & 0.638 & -0.401 & 0.676 \\
Moirai 2.0-R Small & 0.573 & 0.496 & 0.444 & 1.137 & 1.099 & 1.369 & 0.081 & 0.189 & 0.621 & -0.370 & 0.629 \\
TTM R3 & 0.531 & 0.439 & 0.390 & 1.279 & 1.268 & 1.375 & 0.107 & 0.194 & 0.658 & -0.400 & 0.648 \\
TTM R1 & 0.590 & 0.449 & 0.382 & 2.167 & 1.962 & 1.658 & 0.068 & \textbf{0.260} & 0.664 & -0.394 & 0.690 \\
CleanTS-65M & 0.567 & 0.486 & 0.420 & 1.253 & 1.237 & 1.415 & 0.090 & 0.242 & 0.644 & -0.396 & 0.755 \\
TTM R2 & 0.603 & 0.460 & 0.394 & 1.734 & 1.591 & 1.358 & 0.086 & 0.198 & 0.651 & -0.407 & 0.643 \\
YingLong 50M & 0.421 & 0.272 & 0.230 & 1.718 & 1.644 & 1.782 & 0.054 & 0.234 & 0.645 & -0.397 & 0.712 \\
YingLong 300M & 0.419 & 0.269 & 0.227 & 1.724 & 1.654 & 1.779 & 0.053 & 0.254 & 0.642 & -0.398 & 0.755 \\
YingLong 110M & 0.419 & 0.269 & 0.229 & 1.718 & 1.647 & 1.774 & 0.046 & 0.249 & 0.640 & -0.398 & 0.753 \\
YingLong 6M & 0.421 & 0.271 & 0.229 & 1.702 & 1.622 & 1.772 & 0.052 & 0.241 & 0.631 & -0.402 & 0.717 \\
\bottomrule
\end{tabular}
}
\end{table}

\subsection{Performance Across Categories and Evaluation Metrics}

Tables~\ref{tab:categorywise_model_performance} and~\ref{tab:metricwise_model_performance} further reveal substantial heterogeneity in model performance across both financial time-series categories and evaluation metrics. As shown in Table~\ref{tab:categorywise_model_performance}, at the category level, no single model consistently dominates: Chronos-2 Synth performs best for Commodity, Crypto, and Equity, TempoPFN performs best for Foreign Exchange and Market Index, and Toto 2.0 2.5B leads Market Indicators and ETF, while Reverso Small, TiRex 1.1, and VisionTS lead Fixed Income, Macro Indicators, and Fundamentals, respectively. A similar pattern emerges at the metric level, where different models achieve the strongest results for different forecasting objectives. Reverso Small and PatchFM jointly attain the highest hit ratio, while Reverso Small achieves the lowest \(\mathrm{RelMAE}_{ori}\). TiRex 1.1 leads \(\Delta\mathrm{HR}\), \(\Delta^2\mathrm{HR}\), and achieves the lowest \(\mathrm{RelMAE}_{rel}\), while TiRex 1.1 and TiRex 2 Pretrain jointly achieve the lowest \(\mathrm{RelMAE}_{abs}\). For the decision-oriented metrics, Chronos-2 Synth achieves the strongest IC, while TTM R1, Kairos-10M, TimesFM 2.0 500M, and Moirai 1.1-R Large attain the best return, Sharpe ratio, maximum drawdown, and volatility, respectively. Together, these results show that model superiority is highly dependent on both the type of financial time series being forecast and the criterion used to define a useful forecast. FinVerse therefore provides a more informative basis for model selection by allowing practitioners to identify models that are better suited to the characteristics of their target series and decision objectives.

\section{Conclusion}
\label{sec:conclusion}

We introduced FinVerse, a large-scale financial time-series benchmark designed to evaluate forecasting models according to the economic semantics. Moving beyond existing benchmarks that apply uniform conventional forecasting metrics across heterogeneous time series, FinVerse evaluates forecasts through three complementary aspects—point-wise forecasting, cross-sectional ranking, and portfolio backtesting—using evaluation criteria tailored to the economic roles of the target series. Our evaluation demonstrates that performance on generic time-series benchmarks does not necessarily translate to performance under financially informed evaluation. Furthermore, by showing that relative model performance varies substantially across both target financial time series and evaluation metrics, FinVerse provides a more informative basis for model selection according to the economic characteristics of the target series and the intended forecasting objectives. For example, Reverso Small ranks first on FinVerse despite performing less strongly on the generic time-series benchmark, whereas Chronos-2 Synth ranks second on FinVerse while maintaining relatively strong performance across both financial and generic time-series evaluations. We hope that FinVerse provides a foundation for the development and evaluation of future time-series foundation models that better account for the distinct characteristics and decision-relevant properties of financial time series.

\textbf{Limitation \& Future Works.}
FinVerse has several limitations. First, our portfolio backtesting setup largely follows the protocol of \citet{pederzoli2026finbench} for a simple evaluation setting. However, this setup does not fully reflect practical considerations such as transaction costs and turnover. Second, ensuring strict point-in-time (PIT) consistency remains challenging. In particular, the timestamp associated with an observation may differ from its actual release time. Finally, the current benchmark has substantially greater coverage of the U.S. market than other regions. Addressing these limitations remains important future work. Additionally,  we plan to extend FinVerse by incorporating multimodal financial information through methodologies such as~\cite{lee2026fintextsfinancialtextpairedtimeseries}.

\section*{AI Use Statement}
In this work, we used generative AI tools to assist with many parts of the research process. Specifically, we used such tools for implementing methods, cleaning and reformatting data, developing theoretical models or conceptual frameworks, providing feedback on the evaluation methodology, assisting with translation, and drafting interpretations of results. We note that these tools were used only as an aid and only to produce drafts: every AI-assisted output was manually reviewed and verified by the authors before being used. We have not used generative AI tools to generate synthetic datasets, to formulate mathematical claims, to provide ingredients for or assist in the writing of proofs, or to propose or refine hypotheses, and supporting qualitative and thematic data analysis is not applicable to this work. Additionally, we used generative AI tools for tasks with recommended disclosure, including editing code, drafting and editing parts of the manuscript, creating result figures and tables, and summarizing related literature. We take responsibility for the final content of this work, including text, claims, or artifacts produced with the aid of generative AI.

\section*{Ethics Statement}
FinVerse is constructed primarily from publicly accessible financial and economic data sources, such as FRED and the U.S. EIA. During data collection, we made reasonable efforts to review and comply with the applicable licenses, terms of use, and redistribution conditions of the respective data sources. Based on our review, we believe that the data included in the public release can be appropriately used and distributed for research purposes. Nevertheless, given the scale and diversity of the collected data, if any licensing, redistribution, or other compliance issue is identified after release, we will promptly address it by removing, replacing, or otherwise modifying the affected data as appropriate. We are committed to maintaining the dataset in accordance with applicable data-use requirements and to addressing any concerns raised by data providers.

\section*{Reproducibility Statement}
To support reproducibility, we provide the relevant code in the Supplementary Material. Due to the large size of the dataset, we are unable to include the dataset itself in the submission materials. The dataset will be made publicly available upon acceptance.

% FinVerse is constructed primarily from publicly accessible financial and economic data sources, including FRED and the U.S. Energy Information Administration (EIA). During data collection, we made reasonable efforts to review and comply with the applicable licenses, terms of use, and redistribution conditions of the respective data sources. Based on our review, we believe that the data included in the public release can be appropriately used and distributed for research purposes. Nevertheless, given the scale and diversity of the collected data, if any licensing, redistribution, or other compliance issue is identified after release, we will promptly address it by removing, replacing, or otherwise modifying the affected data as appropriate. We are committed to maintaining the dataset in accordance with applicable data-use requirements and to addressing any concerns raised by data providers.

\bibliography{refs}
\bibliographystyle{iclr2027_conference}

% \newpage

\appendix
\section{Detailed Data Coverage}
\label{app:data_coverage}

This appendix provides the detailed coverage table for the released FinVerse data. Table~\ref{tab:appendix_detail_coverage} reports, for each scope, category, detail category, and frequency, whether the group contains evaluated target series, together with the number of series, observed data points in the full data artifact, and valid evaluation points. 

% Two conventions are worth noting when comparing this table with the counts in Section~\ref{sec:data}. The Series and Points columns describe the full data artifact, so they include the history that precedes the evaluation period and the 904 series that are marked here as evaluation targets but whose observations end before it; the 60,232 series and 13.6M observations reported in Section~\ref{sec:data} are the subset that falls inside the evaluation period. The Eval. Points column counts valid forecast-origin cells after filtering rather than raw observations, which is why its total (12.9M) is smaller again.

\begingroup
\setlength{\LTcapwidth}{\textwidth}
\captionsetup[longtable]{font=normalsize}
\scriptsize
\setlength{\tabcolsep}{4pt}
\begin{longtable}{lllllrrr}
\caption{Coverage of the FinVerse data by detail category.}
\label{tab:appendix_detail_coverage}\\
\toprule
Scope & Category & Detail Category & Frequency & Eval. & Series & Points & Eval. Points \\
\midrule
\endfirsthead
% \caption[]{Detailed full-data coverage by scope, category, detail category, frequency, evaluation-target status, observed points, and valid evaluation points.}\\
\toprule
Scope & Category & Detail Category & Frequency & Eval. & Series & Points & Eval. Points \\
\midrule
\endhead
\midrule
\multicolumn{8}{r}{Continued on next page}\\
\endfoot
\bottomrule
\endlastfoot
Inter-country & Foreign Exchange & FX Index & Business-daily & \xmark & 4 & 30,316 & 0 \\
Inter-country & Foreign Exchange & FX Pair & Business-daily & \cmark & 21 & 264,091 & 33,033 \\
Inter-country & Commodity & Commodity ETF Broad & Business-daily & \cmark & 2 & 10,433 & 3,174 \\
Inter-country & Commodity & Commodity ETF Single & Business-daily & \cmark & 5 & 25,154 & 7,935 \\
Inter-country & Commodity & Commodity ETF Thematic & Business-daily & \cmark & 2 & 8,149 & 3,174 \\
Inter-country & Commodity & Future Front & Business-daily & \cmark & 27 & 173,716 & 42,258 \\
Inter-country & Commodity & Inventory Stock & Weekly & \cmark & 7 & 10,985 & 2,303 \\
Inter-country & Commodity & Spot & Business-daily & \cmark & 3 & 28,307 & 4,741 \\
Inter-country & Commodity & Spot & Monthly & \cmark & 13 & 5,123 & 975 \\
Inter-country & Crypto & Crypto Pair Close (Other) & Daily & \xmark & 831 & 794,412 & 0 \\
Inter-country & Crypto & Crypto Pair Close (Top 100) & Daily & \cmark & 100 & 179,443 & 140,483 \\
Inter-country & Crypto & Crypto Pair High & Daily & \xmark & 932 & 959,788 & 0 \\
Inter-country & Crypto & Crypto Pair Low & Daily & \xmark & 931 & 959,424 & 0 \\
Inter-country & Crypto & Crypto Pair Open & Daily & \xmark & 931 & 968,233 & 0 \\
Inter-country & Crypto & Crypto Pair TVL & Daily & \xmark & 24 & 33,254 & 0 \\
Inter-country & Crypto & Crypto Pair Volume & Daily & \xmark & 951 & 1,082,920 & 0 \\
Inter-country & Global Liquidity & FX Year Over Year Change & Quarterly & \xmark & 6 & 801 & 0 \\
Inter-country & Global Liquidity & Stock & Quarterly & \xmark & 15 & 1,692 & 0 \\
Inter-country & Global Liquidity & Year Over Year Change & Quarterly & \xmark & 18 & 1,965 & 0 \\
Inter-country & Global Risk & Credit Spread & Business-daily & \xmark & 1 & 781 & 0 \\
Inter-country & Global Risk & Leading Indicator & Monthly & \xmark & 2 & 1,489 & 0 \\
Inter-country & Global Risk & Uncertainty & Daily & \xmark & 1 & 15,092 & 0 \\
Inter-country & Global Risk & Uncertainty & Monthly & \xmark & 2 & 694 & 0 \\
Inter-country & Market Index & Country Index & Business-daily & \cmark & 20 & 155,667 & 28,422 \\
Inter-country & Market Index & Regional Index & Business-daily & \cmark & 2 & 10,724 & 3,184 \\
Country & Fixed Income & Bond Total Return Index & Business-daily & \cmark & 10 & 7,830 & 7,750 \\
Country & Fixed Income & Breakeven & Business-daily & \cmark & 2 & 12,172 & 3,166 \\
Country & Fixed Income & Breakeven & Monthly & \cmark & 2 & 455 & 150 \\
Country & Fixed Income & Corporate Yield & Business-daily & \cmark & 2 & 21,823 & 3,160 \\
Country & Fixed Income & Corporate Yield & Monthly & \cmark & 2 & 2,574 & 150 \\
Country & Fixed Income & Credit Spread & Business-daily & \cmark & 10 & 7,830 & 7,749 \\
Country & Fixed Income & Forward Inflation & Business-daily & \cmark & 1 & 6,086 & 1,583 \\
Country & Fixed Income & Inflation Expectation & Monthly & \cmark & 7 & 3,949 & 530 \\
Country & Fixed Income & Money Market & Business-daily & \cmark & 8 & 20,958 & 12,317 \\
Country & Fixed Income & Mortgage Rate & Weekly & \cmark & 3 & 5,617 & 812 \\
Country & Fixed Income & Swap Rate & Business-daily & \xmark & 7 & 29,820 & 0 \\
Country & Fixed Income & TIPS Yield & Business-daily & \cmark & 4 & 22,478 & 6,328 \\
Country & Fixed Income & Term Spread & Business-daily & \cmark & 2 & 24,587 & 3,166 \\
Country & Fixed Income & Treasury Bill & Business-daily & \xmark & 4 & 60,333 & 0 \\
Country & Fixed Income & Treasury Yield & Business-daily & \cmark & 11 & 154,357 & 17,402 \\
Country & Market Indicators & Fin Conditions & Weekly & \xmark & 6 & 17,316 & 0 \\
Country & Market Indicators & Fin Stress & Weekly & \xmark & 1 & 1,687 & 0 \\
Country & Market Indicators & Fin Stress & Monthly & \xmark & 1 & 434 & 0 \\
Country & Market Indicators & Implied Volatility & Business-daily & \cmark & 9 & 55,495 & 12,879 \\
Country & Market Indicators & Leading Index & Monthly & \xmark & 1 & 794 & 0 \\
Country & Market Indicators & Money Spread & Business-daily & \xmark & 1 & 9,407 & 0 \\
Country & Market Indicators & Recession Flag & Daily & \xmark & 1 & 62,606 & 0 \\
Country & Market Indicators & Recession Flag & Monthly & \xmark & 1 & 2,056 & 0 \\
Country & Market Indicators & Recession Prob & Monthly & \xmark & 1 & 705 & 0 \\
Country & Market Indicators & Sentiment & Business-daily & \xmark & 1 & 9,478 & 0 \\
Country & Market Indicators & US Equity Index & Business-daily & \cmark & 9 & 110,839 & 14,310 \\
Country & Macro Indicators & Activity & Monthly & \cmark & 10 & 5,621 & 743 \\
Country & Macro Indicators & Banking & Weekly & \xmark & 2 & 5,562 & 0 \\
Country & Macro Indicators & Banking & Monthly & \cmark & 2 & 1,902 & 150 \\
Country & Macro Indicators & Capacity Utilization By Industry & Monthly & \cmark & 11 & 8,385 & 825 \\
Country & Macro Indicators & Credit & Monthly & \cmark & 2 & 1,696 & 148 \\
Country & Macro Indicators & Current Account & Quarterly & \cmark & 4 & 650 & 48 \\
Country & Macro Indicators & Energy Activity & Monthly & \cmark & 6 & 2,557 & 397 \\
Country & Macro Indicators & Energy Refinery & Weekly & \xmark & 5 & 7,302 & 0 \\
Country & Macro Indicators & External & Monthly & \cmark & 6 & 2,460 & 444 \\
Country & Macro Indicators & External & Quarterly & \cmark & 3 & 951 & 75 \\
Country & Macro Indicators & FDI Flows & Quarterly & \cmark & 1 & 317 & 24 \\
Country & Macro Indicators & FDI Flows & Annual & \cmark & 1 & 80 & 6 \\
Country & Macro Indicators & Financial Flows & Quarterly & \cmark & 1 & 316 & 24 \\
Country & Macro Indicators & Fiscal & Monthly & \cmark & 1 & 546 & 75 \\
Country & Macro Indicators & Fiscal & Quarterly & \cmark & 1 & 240 & 24 \\
Country & Macro Indicators & Fiscal & Annual & \cmark & 2 & 222 & 12 \\
Country & Macro Indicators & Freight & Monthly & \cmark & 1 & 313 & 73 \\
Country & Macro Indicators & Gdp & Quarterly & \cmark & 2 & 634 & 50 \\
Country & Macro Indicators & Housing & Monthly & \cmark & 4 & 2,367 & 231 \\
Country & Macro Indicators & Housing Price & Monthly & \cmark & 10 & 3,099 & 747 \\
Country & Macro Indicators & Housing Price & Quarterly & \xmark & 1 & 204 & 0 \\
Country & Macro Indicators & Income & Monthly & \cmark & 3 & 2,421 & 225 \\
Country & Macro Indicators & Industrial Production By Industry & Monthly & \cmark & 28 & 19,776 & 2,100 \\
Country & Macro Indicators & Inflation & Monthly & \cmark & 10 & 6,793 & 673 \\
Country & Macro Indicators & International Investment Position & Quarterly & \cmark & 5 & 400 & 120 \\
Country & Macro Indicators & International Investment Position & Annual & \cmark & 1 & 50 & 6 \\
Country & Macro Indicators & Inventory & Monthly & \cmark & 7 & 2,870 & 518 \\
Country & Macro Indicators & Labor & Weekly & \xmark & 2 & 6,189 & 0 \\
Country & Macro Indicators & Labor & Monthly & \cmark & 14 & 7,872 & 1,041 \\
Country & Macro Indicators & Labor By Industry & Monthly & \cmark & 16 & 15,744 & 1,200 \\
Country & Macro Indicators & Money & Weekly & \xmark & 1 & 1,220 & 0 \\
Country & Macro Indicators & Money & Monthly & \cmark & 5 & 4,035 & 374 \\
Country & Macro Indicators & Policy Rate & Business-daily & \xmark & 1 & 18,453 & 0 \\
Country & Macro Indicators & Policy Rate & Daily & \xmark & 3 & 22,267 & 0 \\
Country & Macro Indicators & Portfolio Flows & Quarterly & \cmark & 7 & 2,231 & 168 \\
Country & Macro Indicators & Productivity & Quarterly & \cmark & 3 & 948 & 72 \\
Country & Macro Indicators & Regional Fed & Monthly & \cmark & 2 & 994 & 152 \\
Country & Macro Indicators & Saving & Monthly & \cmark & 1 & 807 & 75 \\
Country & Macro Indicators & Sentiment & Monthly & \cmark & 3 & 1,763 & 200 \\
Country & Macro Indicators & Spending & Monthly & \cmark & 6 & 3,690 & 450 \\
Country & Macro Indicators & Stress & Monthly & \cmark & 1 & 434 & 75 \\
Country & Macro Indicators & TIC Flows & Quarterly & \cmark & 4 & 851 & 95 \\
Individual & ETF & ETF OHLCV Close & Business-daily & \cmark & 5,107 & 8,100,873 & 4,497,246 \\
Individual & ETF & ETF OHLCV High & Business-daily & \xmark & 5,107 & 8,098,720 & 0 \\
Individual & ETF & ETF OHLCV Low & Business-daily & \xmark & 5,107 & 8,095,707 & 0 \\
Individual & ETF & ETF OHLCV Open & Business-daily & \xmark & 5,107 & 8,108,763 & 0 \\
Individual & ETF & ETF OHLCV Volume & Business-daily & \xmark & 5,103 & 8,143,493 & 0 \\
Individual & Equity & Equity OHLCV Close & Business-daily & \cmark & 6,143 & 24,193,386 & 7,963,766 \\
Individual & Equity & Equity OHLCV High & Business-daily & \xmark & 6,144 & 24,189,072 & 0 \\
Individual & Equity & Equity OHLCV Low & Business-daily & \xmark & 6,144 & 24,106,760 & 0 \\
Individual & Equity & Equity OHLCV Open & Business-daily & \xmark & 6,145 & 24,253,740 & 0 \\
Individual & Equity & Equity OHLCV Volume & Business-daily & \xmark & 6,169 & 24,986,230 & 0 \\
Individual & Fundamentals & Capital Expenditures & Quarterly & \cmark & 3,442 & 140,619 & 48,017 \\
Individual & Fundamentals & Cash and Cash Equivalents & Quarterly & \cmark & 4,642 & 206,110 & 81,457 \\
Individual & Fundamentals & EPS Basic & Quarterly & \cmark & 4,185 & 169,330 & 62,002 \\
Individual & Fundamentals & EPS Diluted & Quarterly & \cmark & 4,166 & 166,806 & 60,690 \\
Individual & Fundamentals & Long-Term Debt & Quarterly & \cmark & 3,249 & 100,936 & 40,488 \\
Individual & Fundamentals & Net Income & Quarterly & \cmark & 4,357 & 182,999 & 65,622 \\
Individual & Fundamentals & Operating Cash Flow & Quarterly & \cmark & 4,359 & 185,546 & 67,429 \\
Individual & Fundamentals & Operating Income & Quarterly & \cmark & 3,707 & 151,714 & 54,499 \\
Individual & Fundamentals & R And D Expense & Quarterly & \xmark & 1,851 & 64,374 & 0 \\
Individual & Fundamentals & Revenue & Quarterly & \cmark & 3,677 & 99,475 & 49,241 \\
Individual & Fundamentals & Shares Diluted & Quarterly & \xmark & 4,173 & 168,275 & 0 \\
Individual & Fundamentals & Stockholders Equity & Quarterly & \cmark & 4,697 & 210,564 & 90,538 \\
Individual & Fundamentals & Total Assets & Quarterly & \cmark & 4,699 & 196,491 & 90,137 \\
Individual & Fundamentals & Total Liabilities & Quarterly & \cmark & 4,238 & 160,730 & 77,290 \\
Individual & News Sentiment & Sentiment Tone & Daily & \xmark & 11 & 37,488 & 0 \\
Individual & News Sentiment & Sentiment Volume & Daily & \xmark & 11 & 37,488 & 0 \\
\end{longtable}
\endgroup

%=====================================================================
% Appendix B. Metric Assignment by Data Category
%=====================================================================
\section{Foundation Model List}
\label{app:foundation_list}
The evaluated models include TimesFM 1.0 200M, TimesFM 2.0 500M, and TimesFM 2.5 200M~\citep{das2024decoder}; Chronos-2, Chronos-2 Synth, Chronos-Bolt Small, and Chronos-Bolt Base~\citep{ansari2024chronos,ansari2025chronos2}; Moirai 1.1-R Small, Moirai 1.1-R Base, Moirai 1.1-R Large, and Moirai 2.0-R Small~\citep{woo2024unified}; Timer-S1~\citep{liu2024timer,liu2025timerxl}; TempoPFN~\citep{moroshan2026tempopfnsyntheticpretraininglinear}; Toto Open Base 1.0, Toto 2.0 4M, Toto 2.0 22M, Toto 2.0 313M, Toto 2.0 1B, and Toto 2.0 2.5B~\citep{toto2025}; TiRex 1.1, TiRex 2 Pretrain, and TiRex 2 ZS~\citep{dooley2025tirex}; VisionTS~\citep{chen2024visionts}; Sundial Base 128M~\citep{liu2025sundial}; YingLong 6M, YingLong 50M, YingLong 110M, and YingLong 300M~\citep{li2025yinglong}; Kairos-10M, Kairos-23M, and Kairos-50M~\citep{kairos2025}; Reverso Small~\citep{reverso2025}; TTM R1, TTM R2, and TTM R3~\citep{ekambaram2024tinytimemixer}; PatchTST-FM R1 and Granite PatchTST-FM R1~\citep{nie2023patchtst}; FlowState and Granite FlowState R1~\citep{ibm2025flowstate}; PatchFM~\citep{patchfm}; T0 Alpha~\citep{t0alpha}; Super Linear~\citep{superlinear}; and CleanTS-65M~\citep{cleants65m}.  

\section{Metric Definition}
\label{app:metric}

This section provides the detailed mathematical formulations of the evaluation metrics used in FinVerse. For a single series, let $y_t$ denote the realized value at time $t$, let $\hat{y}_{t+h}$ denote the forecast for horizon $h$, and let $\mathcal{O}$ be the set of valid forecast origins. Throughout, a forecast origin $o \in \mathcal{O}$ denotes the last time step observed by the model, so that the forecast window covers $o+1,\ldots,o+H$ and $y_o$ is the most recent observation available when the forecast is made. 

% Consequently, every baseline used below is constructed only from observations up to and including $y_o$.

\subsection{Hit Ratio}
Hit ratio measures directional correctness. For a comparison window $p$ and forecast horizon $h$, we define the realized and predicted changes as
\begin{equation}
r_o^{(p,h)}
=
\frac{y_{o+h}}{y_{o+h-p}} - 1,
\qquad
\hat{r}_o^{(p,h)}
=
\frac{\hat{y}_{o+h}}{y_{o+h-p}} - 1.
\end{equation}

To capture higher-order directional dynamics, we further define
\begin{align}
\Delta r_o^{(p,h)}
&=
r_o^{(p,h)} - r_{o-1}^{(p,h)},
&
\Delta \hat{r}_o^{(p,h)}
&=
\hat{r}_o^{(p,h)} - r_{o-1}^{(p,h)},
\\
\Delta^2 r_o^{(p,h)}
&=
\Delta r_o^{(p,h)} - \Delta r_{o-1}^{(p,h)},
&
\Delta^2 \hat{r}_o^{(p,h)}
&=
\Delta \hat{r}_o^{(p,h)} - \Delta r_{o-1}^{(p,h)}.
\end{align}

Let
$\Delta^0 r_o^{(p,h)} = r_o^{(p,h)}$
and
$\Delta^0 \hat{r}_o^{(p,h)} = \hat{r}_o^{(p,h)}$.
We define the $k$-th order hit ratio as
\begin{equation}
\Delta^k \mathrm{HR}^{(p,h)}
=
\frac{1}{|\mathcal{O}|}
\sum_{o \in \mathcal{O}}
\prod_{j=0}^{k}
\mathbb{I}\!\left[
\operatorname{sign}\!\left(
\Delta^j \hat{r}_o^{(p,h)}
\right)
=
\operatorname{sign}\!\left(
\Delta^j r_o^{(p,h)}
\right)
\right],
\qquad
k \in \{0,1,2\}.
\end{equation}

Here, $k=0$, $k=1$, and $k=2$ correspond to
$\mathrm{HR}$, $\Delta\mathrm{HR}$, and
$\Delta^2\mathrm{HR}$, respectively. The comparison window \(p\) specifies the reference interval used to compute the realized and predicted changes (e.g., week-over-week, month-over-month, quarter-over-quarter, half-year-over-half-year, or year-over-year), whereas the forecast horizon \(h\) specifies how far into the future the prediction is evaluated. For example, \(p=\mathrm{1y}\) computes the year-over-year change relative to the observation one year earlier, while \(h=\mathrm{1y}\) evaluates the prediction one year ahead. Different financial series use different $(p,h)$ combinations according to their sampling frequency and economic interpretation. Diagonal variants such as \((1m,1m)\) or \((1y,1y)\) measure the change from the forecast origin to the target endpoint, which is the natural setting for asset-like returns, cross-sectional ranking, and portfolio horizons. Non-diagonal variants such as \((1y,1m)\) measure a longer-window quantity at a shorter forecast horizon; for example, they evaluate the year-over-year value that will be observed one month after the forecast origin.

% The complete $(p,h)$ assignments for each financial series and evaluation metric are provided in Appendix Tables~\ref{tab:appendix_metric_assignment_intercountry_foreign_exchange}--\ref{tab:appendix_metric_assignment_individual_fundamentals}.

\subsection{Relative Mean Absolute Error}
FinVerse uses \(\mathrm{RelMAE}\), defined as the mean absolute error of the forecast divided by the mean absolute error of a naive forecast computed over the same evaluation cells. \(\mathrm{RelMAE}_{rel}\) evaluates relative-change forecasts against a zero-change baseline. In other words, the baseline assumes no relative change over the forecast horizon:
\begin{equation}
    \mathrm{RelMAE}_{rel}^{(p,h)} =
    \frac{
    \frac{1}{|\mathcal{O}|}
    \sum_{o \in \mathcal{O}}
    \left|\hat{r}_{o}^{(p,h)} - r_{o}^{(p,h)}\right|
    }{
    \frac{1}{|\mathcal{O}|}
    \sum_{o \in \mathcal{O}}
    \left|r_{o}^{(p,h)}\right|
    }.
\end{equation}

\(\mathrm{RelMAE}_{abs}\) evaluates raw-value forecasts against a no-change raw-value baseline defined over the same comparison window. In other words, the baseline predicts the future value by repeating the observation at \(y_{o+h-p}\):
\begin{equation}
    \mathrm{RelMAE}_{abs}^{(p,h)} =
    \frac{
    \frac{1}{|\mathcal{O}|}
    \sum_{o \in \mathcal{O}}
    \left|\hat{y}_{o+h} - y_{o+h}\right|
    }{
    \frac{1}{|\mathcal{O}|}
    \sum_{o \in \mathcal{O}}
    \left|y_{o+h} - y_{o+h-p}\right|
    }.
\end{equation}

Finally, \(\mathrm{RelMAE}_{ori}\) evaluates raw-value forecasts against the conventional origin-value persistence baseline, where the most recently observed value is repeated for all future horizons. This baseline is the persistence forecast from the origin, and the metric averages over every horizon \(h=1,\dots,H\) rather than over a selected \((p,h)\) grid:
\begin{equation}
    \mathrm{RelMAE}_{ori}
    =
    \frac{
    \frac{1}{|\mathcal{O}|H}
    \sum_{o \in \mathcal{O}}
    \sum_{h=1}^{H}
    \left|\hat{y}_{o+h} - y_{o+h}\right|
    }{
    \frac{1}{|\mathcal{O}|H}
    \sum_{o \in \mathcal{O}}
    \sum_{h=1}^{H}
    \left|y_o - y_{o+h}\right|
    }.
\end{equation}

\subsection{Information Coefficient}
The information coefficient evaluates cross-sectional ranking quality. For a set of assets $\mathcal{S}_o$ available at origin $o$, FinVerse computes a Spearman rank correlation between predicted and realized future changes:
\begin{equation}
    \mathrm{IC}_{o}^{(p,h)}
    =
    \rho_{\mathrm{Spearman}}
    \left(
    \left\{\hat{r}_{i,o}^{(p,h)}\right\}_{i \in \mathcal{S}_o},
    \left\{r_{i,o}^{(p,h)}\right\}_{i \in \mathcal{S}_o}
    \right).
\end{equation}

% Origins at which fewer than three assets have both a finite prediction and a finite realization are discarded, and ties are ranked by their average rank. The value reported for a group is the mean of \(\mathrm{IC}_{o}^{(p,h)}\) over origins weighted by the number of assets entering each origin, so that origins with a wider cross-section, and hence a more reliable rank correlation, count proportionally more.

\subsection{Portfolio Backtesting}
Portfolio evaluation converts forecasts into an equal-weight long-only strategy. At each rebalance origin, the benchmark ranks assets by the horizon-\(h\) predicted signal and selects the top decile, subject to a floor of ten instruments: \(|\mathcal{P}_o^{(h)}| = \min(|\mathcal{S}_o|, \max(10, \lceil 0.1\,|\mathcal{S}_o|\rceil))\).   For a holding horizon \(h\), the selected set \(\mathcal{P}_o^{(h)}\) is reconstructed at rebalance origins spaced \(h\) frequency steps apart and held for the next \(h\) steps. To avoid depending on a single calendar start date, FinVerse runs this backtest for every possible offset within the \(h\)-step window and averages the resulting portfolio statistics. Thus, for a one-month horizon, starts such as January 1 and January 2 define separate monthly-rebalanced paths rather than discarding all but one start date. The realized portfolio return for a rebalance origin is
\begin{equation}
    R_o^{(h)}
    =
    \frac{1}{|\mathcal{P}_o^{(h)}|}
    \sum_{i \in \mathcal{P}_o^{(h)}}
    r_{i,o}^{(h,h)}.
\end{equation}
Let \(\mathcal{A}_h=\{0,\ldots,h-1\}\) denote the offset set, and let \(o_{s,1},\ldots,o_{s,K_s}\) denote the rebalance origins in chronological order for offset \(s\), where \(K_s\) is the number of valid rebalancing periods in that offset path. The cumulative net asset value for offset \(s\) is
\begin{equation}
    \mathrm{NAV}_{s,k}^{(h)}
    =
    \prod_{j=1}^{k}
    \left(1 + R_{o_{s,j}}^{(h)}\right).
\end{equation}
Because \(\mathrm{NAV}_{s,k}^{(h)}\) is observed only at rebalancing dates, it cannot reveal losses that occur and recover within a holding period. FinVerse therefore also tracks the portfolio value at every frequency step. Within the \(j\)-th holding period, the selected assets are bought equal-weight at \(y_{i,o_{s,j}}\), the most recent observation available at that origin, and held, so the portfolio value after \(m\) steps is
\begin{equation}
    V_{s,(j,m)}^{(h)}
    =
    \mathrm{NAV}_{s,j-1}^{(h)}
    \cdot
    \frac{1}{|\mathcal{P}_{o_{s,j}}^{(h)}|}
    \sum_{i \in \mathcal{P}_{o_{s,j}}^{(h)}}
    \frac{y_{i,\,o_{s,j}+m}}{y_{i,\,o_{s,j}}},
    \qquad m = 1,\ldots,h,
\end{equation}
with \(\mathrm{NAV}_{s,0}^{(h)}=1\). Because the holding is buy-and-hold, the end of each period coincides with the rebalance-level value, \(V_{s,(j,h)}^{(h)}=\mathrm{NAV}_{s,j}^{(h)}\). Concatenating these values over \(j=1,\ldots,K_s\) yields the step-level value path \(V_{s,1}^{(h)},\ldots,V_{s,T_s}^{(h)}\) with \(T_s=K_s h\), and the corresponding step returns are
\begin{equation}
    R^{\mathrm{step}}_{s,u}
    =
    \frac{V_{s,u}^{(h)}}{V_{s,u-1}^{(h)}} - 1,
    \qquad V_{s,0}^{(h)} = 1 .
\end{equation}

The resulting sequences are summarized with annualized return, annualized volatility, Sharpe ratio, and maximum drawdown for each offset, and the reported metric is the average across offsets:
\begin{equation}
    \mathrm{Metric}^{(h)}
    =
    \frac{1}{|\mathcal{A}_h|}
    \sum_{s \in \mathcal{A}_h}
    \mathrm{Metric}_{s}^{(h)}.
\end{equation}
Here, \(h\) denotes the holding horizon. Let \(P_f\) denote the number of frequency steps per year, with \(P_f \in \{252,365,52,12,4,1\}\) for business-daily, daily, weekly, monthly, quarterly, and annual series, respectively. The number of rebalancing trades per year is then \(A_h = P_f / h\). For each offset \(s\), the annualized return is computed from the rebalance-level returns,
\begin{equation}
    \mathrm{AnnualReturn}_{s}^{(h)}
    =
    \left(
    \prod_{j=1}^{K_s}
    \left(1+R_{o_{s,j}}^{(h)}\right)
    \right)^{\frac{A_h}{K_s}}
    -1,
\end{equation}
whereas the risk metrics are computed from the step-level path, so that losses occurring inside a holding period are reflected:
\begin{equation}
    \mathrm{AnnualVolatility}_{s}^{(h)}
    =
    \mathrm{Std}\!\left(R^{\mathrm{step}}_{s,u}\right)
    \sqrt{P_f},
    \qquad
    \mathrm{Sharpe}_{s}^{(h)}
    =
    \frac{\mathbb{E}\!\left[R^{\mathrm{step}}_{s,u}\right]}{\mathrm{Std}\!\left(R^{\mathrm{step}}_{s,u}\right)}
    \sqrt{P_f},
\end{equation}
\begin{equation}
    \mathrm{MDD}_{s}^{(h)}
    =
    \min_{u}
    \left(
    \frac{V_{s,u}^{(h)}}{\max\!\left(1,\ \max_{v \le u} V_{s,v}^{(h)}\right)} - 1
    \right).
\end{equation}
The peak in the drawdown denominator includes the initial value of one, so a decline that starts before the first rebalancing date is also counted.

\subsection{Return, Volatility, and Sharpe Ratio Aggregation}
\label{app:reason_sharpe}
The Sharpe column is not the ratio of the Return and Vol. columns for three reasons. First, even within a single backtest, annualized return and the numerator of the Sharpe ratio are computed differently: annualized return is based on the compounded growth of the portfolio net asset value, whereas the Sharpe ratio uses the annualized arithmetic mean of one-step portfolio returns. Second, these statistics are computed separately for each holding-period offset and then averaged across offsets. As such, the average Sharpe ratio need not equal the ratio of the average return to the average volatility. Third, each column in the table is independently aggregated using a mean across all evaluated detail category–frequency units and holding horizons, so the ratio of the aggregated Return and Vol. columns need not equal the aggregated Sharpe ratio. 
% For Reverso Small, for example, the reported columns give \(0.171/0.470=0.37\), whereas the reported Sharpe ratio is \(0.697\).

\subsection{Assumptions of Portfolio Evaluation}
\label{app:assume_port}
% The backtest is deliberately simple, and its numbers should be read with the following assumptions in mind.

\begin{enumerate}
\item \textbf{Trading frictions.} No transaction costs, bid--ask spreads, or market impact are charged. Every rebalancing date replaces the entire portfolio with the newly selected instruments, so turnover is effectively complete at each rebalance and is higher for the shorter holding horizons.

\item \textbf{Investable set, listing, and liquidity.} At each rebalancing date the portfolio is formed from the instruments that have an observed price at that date, and the availability of a price is taken as evidence that the instrument was listed and tradable then. An instrument whose price series ends simply leaves the cross-section from that date onward.

\item \textbf{Corporate actions.} Dividends and other cash distributions are not added back, so equity, ETF, and index returns are price returns rather than total returns, and the reported portfolio performance is correspondingly conservative for dividend-paying assets.

\item \textbf{Portfolio construction.} Positions are entered at the last observed price before the rebalancing origin, weighted equally at entry, and held without intra-period rebalancing, so weights drift with prices inside a holding period. The strategy is long-only, unlevered, and fully invested in the selected instruments. To avoid dependence on a single calendar start date, the backtest is repeated for every offset within the holding horizon and the resulting statistics are averaged.

\item \textbf{Trading calendars and the step-level path.} The step-level value path is evaluated on the union of the one-step date grids of the instruments in the group, and an instrument without an observation at a grid step carries its most recent price forward.

\item \textbf{Non-equity instruments.} The same rule is applied to every group for which a cross-sectional portfolio is meaningful, and each group is traded as its published price series. Foreign-exchange pairs use the quoted reference rate as published, so a position is a long exposure to that quoted rate and interest-rate differentials, i.e., carry, are not modeled. Commodity futures use front-month contract prices, so roll returns between contracts are not modeled. Commodity spot series are reference prices, which ignores storage and financing costs and makes them the least directly investable group. Country equity indices are price indices, so, as with equities and ETFs, dividends are excluded. Crypto pairs are daily closing prices on a seven-day calendar and are treated like any other price series. These groups therefore measure the value of the forecast signal for ranking instruments, not the return of an implementable trading strategy in each market.
\end{enumerate}

\begin{table}[t]
\centering
\caption{Mapping from compact variant IDs to \((p,h)\) combinations. \(1w\), \(1m\), \(1q\), \(1h\), and \(1y\) denote one week, one month, one quarter, one half-year, and one year, respectively.}
\label{tab:appendix_variant_mapping}
\vspace{0.35em}
\scriptsize
\setlength{\tabcolsep}{10pt}
\renewcommand{\arraystretch}{1.08}
\resizebox{\linewidth}{!}{
% \begin{adjustbox}{max width=\linewidth}
\begin{tabular}{ll}
\toprule
\((p,h)\) Variant & \((p,h)\) combinations \\
\midrule
D5  & \((1w,1w),(1m,1m),(1q,1q),(1h,1h),(1y,1y)\) \\
D4  & \((1m,1m),(1q,1q),(1h,1h),(1y,1y)\) \\
M10 & \((1m,1m),(1q,1m),(1q,1q),(1h,1m),(1h,1q),(1h,1h),(1y,1m),(1y,1q),(1y,1h),(1y,1y)\) \\
M4  & \((1m,1m),(1q,1m),(1h,1m),(1y,1m)\) \\
Q6  & \((1q,1q),(1h,1q),(1h,1h),(1y,1q),(1y,1h),(1y,1y)\) \\
Q3  & \((1q,1q),(1h,1q),(1y,1q)\) \\
A1  & \((1y,1y)\) \\
N/A & N/A \\
\bottomrule
\end{tabular}
}
\end{table}

\section{Metric Assignment by Data Category}
\label{app:metric_assignment}

This section shows the metric assignment used by FinVerse. Table~\ref{tab:appendix_variant_mapping} maps each compact variant ID to the corresponding \((p,h)\) combinations. Tables~\ref{tab:appendix_metric_assignment_intercountry_foreign_exchange}--\ref{tab:appendix_metric_assignment_individual_fundamentals} then provide the detailed metric assignment for each evaluated scope-category group. Each table lists the detail category, frequency, number of evaluation points, evaluation aspect, metric families, and \((p,h)\) variant used for that task. 

% \paragraph{Interpretation of \((p,h)\) variants.}
% The notation \((p,h)\) follows the target definition in Section~\ref{subsec:benchmark_evaluation_metrics}. For a forecast origin \(o\), \(h\) is the forecast horizon, so the forecast is evaluated at the future endpoint \(o+h\). The parameter \(p\) is the comparison window ending at that endpoint: relative-change metrics use \(y_{o+h-p}\) as the reference value and evaluate \(r_o^{(p,h)}=y_{o+h}/y_{o+h-p}-1\). Thus, \(p\) determines the economic movement being measured, while \(h\) determines how far ahead that movement is evaluated. Diagonal variants such as \((1m,1m)\) or \((1y,1y)\) measure the change from the forecast origin to the target endpoint, which is the natural setting for asset-like returns, cross-sectional ranking, and portfolio horizons. Non-diagonal variants such as \((1y,1m)\) measure a longer-window quantity at a shorter forecast horizon; for example, they evaluate the year-over-year value that will be observed one month after the forecast origin. In the compact IDs below, \(1w\), \(1m\), \(1q\), \(1h\), and \(1y\) denote one week, one month, one quarter, one half-year, and one year, respectively. The variant IDs are used only to keep the metric-assignment tables compact; the corresponding \((p,h)\) combinations are listed explicitly in Table~\ref{tab:appendix_variant_mapping}.

\FloatBarrier
\begingroup
\captionsetup[table]{font=normalsize}
\setlength{\LTcapwidth}{\textwidth}
\setlength{\tabcolsep}{6pt}
\renewcommand{\arraystretch}{1.04}
{\fontsize{8.0pt}{9pt}\selectfont
\begin{longtable}{@{}llllll@{}}
\caption{Metric assignments and variant IDs for inter-country Foreign Exchange  categories.}
\label{tab:appendix_metric_assignment_intercountry_foreign_exchange}\\
\toprule
Detail Category & Frequency & Eval. Points & Task & Metrics & \((p,h)\) Variant \\
\midrule
\endfirsthead
\toprule
Detail Category & Frequency & Eval. Points & Task & Metrics & \((p,h)\) Variant \\
\midrule
\endhead
\midrule
\multicolumn{6}{r}{\emph{Continued on next page}} \\
\endfoot
\bottomrule
\endlastfoot
\multirow{4}{*}{FX Pair} & \multirow{4}{*}{Business-daily} & \multirow{4}{*}{33,033} & Point-wise & HR, \(\mathrm{RelMAE}_{rel}\), \(\mathrm{RelMAE}_{abs}\) & D5 \\
 &  &  & Point-wise & \(\mathrm{RelMAE}_{ori}\) & N/A \\
 &  &  & Cross-sectional & IC & D5 \\
 &  &  & Portfolio & \mbox{return, vol, Sharpe, MDD} & D5 \\
\end{longtable}
}
\endgroup

\begingroup
\tiny
\captionsetup[table]{font=normalsize}
\setlength{\LTcapwidth}{\textwidth}
\setlength{\tabcolsep}{2.7pt}
\renewcommand{\arraystretch}{1.15}
{\fontsize{8.0pt}{9pt}\selectfont
\begin{longtable}{@{}llllll@{}}
\caption{Metric assignments and variant IDs for inter-country Commodity categories.}
\label{tab:appendix_metric_assignment_intercountry_commodity}\\
\toprule
Detail Category & Frequency & Eval. Points & Task & Metrics & \((p,h)\) Variant \\
\midrule
\endfirsthead
\toprule
Detail Category & Frequency & Eval. Points & Task & Metrics & \((p,h)\) Variant \\
\midrule
\endhead
\midrule
\multicolumn{6}{r}{\emph{Continued on next page}} \\
\endfoot
\bottomrule
\endlastfoot
\multirow{4}{*}{Future Front} & \multirow{4}{*}{Business-daily} & \multirow{4}{*}{42,258} & Point-wise & HR, \(\mathrm{RelMAE}_{rel}\), \(\mathrm{RelMAE}_{abs}\) & D5 \\
 &  &  & Point-wise & \(\mathrm{RelMAE}_{ori}\) & N/A \\
 &  &  & Cross-sectional & IC & D5 \\
 &  &  & Portfolio & \mbox{return, vol, Sharpe, MDD} & D5 \\
\midrule
\multirow{2}{*}{Spot} & \multirow{2}{*}{Business-daily} & \multirow{2}{*}{4,741} & Point-wise & HR, \(\mathrm{RelMAE}_{rel}\), \(\mathrm{RelMAE}_{abs}\) & D5 \\
 &  &  & Point-wise & \(\mathrm{RelMAE}_{ori}\) & N/A \\
\midrule
\multirow{4}{*}{Spot} & \multirow{4}{*}{Monthly} & \multirow{4}{*}{975} & Point-wise & HR, \(\mathrm{RelMAE}_{rel}\), \(\mathrm{RelMAE}_{abs}\) & D4 \\
 &  &  & Point-wise & \(\mathrm{RelMAE}_{ori}\) & N/A \\
\end{longtable}
}
\endgroup

\begingroup
\captionsetup[table]{font=normalsize}
\setlength{\LTcapwidth}{\textwidth}
\setlength{\tabcolsep}{3.8pt}
\renewcommand{\arraystretch}{1.15}
{\fontsize{8.0pt}{9pt}\selectfont
\begin{longtable}{@{}llllll@{}}
\caption{Metric assignments and variant IDs for inter-country Crypto categories.}
\label{tab:appendix_metric_assignment_intercountry_crypto}\\
\toprule
Detail Category & Frequency & Eval. Points & Task & Metrics & \((p,h)\) Variant \\
\midrule
\endfirsthead
\toprule
Detail Category & Frequency & Eval. Points & Task & Metrics & \((p,h)\) Variant \\
\midrule
\endhead
\midrule
\multicolumn{6}{r}{\emph{Continued on next page}} \\
\endfoot
\bottomrule
\endlastfoot
\multirow{4}{*}{Crypto Pair Close (Top 100)} & \multirow{4}{*}{Daily} & \multirow{4}{*}{140,483} & Point-wise & HR, \(\mathrm{RelMAE}_{rel}\), \(\mathrm{RelMAE}_{abs}\) & D5 \\
 &  &  & Point-wise & \(\mathrm{RelMAE}_{ori}\) & N/A \\
 &  &  & Cross-sectional & IC & D5 \\
 &  &  & Portfolio & \mbox{return, vol, Sharpe, MDD} & D5 \\
\end{longtable}
}
\endgroup

\begingroup
\tiny
\captionsetup[table]{font=normalsize}
\setlength{\LTcapwidth}{\textwidth}
\setlength{\tabcolsep}{6pt}
\renewcommand{\arraystretch}{1.15}
{\fontsize{8.0pt}{9pt}\selectfont
\begin{longtable}{@{}llllll@{}}
\caption{Metric assignments and variant IDs for inter-country Market Index categories.}
\label{tab:appendix_metric_assignment_intercountry_market_index}\\
\toprule
Detail Category & Frequency & Eval. Points & Task & Metrics & \((p,h)\) Variant \\
\midrule
\endfirsthead
\toprule
Detail Category & Frequency & Eval. Points & Task & Metrics & \((p,h)\) Variant \\
\midrule
\endhead
\midrule
\multicolumn{6}{r}{\emph{Continued on next page}} \\
\endfoot
\bottomrule
\endlastfoot
\multirow{4}{*}{Country Index} & \multirow{4}{*}{Business-daily} & \multirow{4}{*}{28,422} & Point-wise & HR, \(\mathrm{RelMAE}_{rel}\), \(\mathrm{RelMAE}_{abs}\) & D5 \\
 &  &  & Point-wise & \(\mathrm{RelMAE}_{ori}\) & N/A \\
 &  &  & Cross-sectional & IC & D5 \\
 &  &  & Portfolio & \mbox{return, vol, Sharpe, MDD} & D5 \\
\midrule
\multirow{2}{*}{Regional Index} & \multirow{2}{*}{Business-daily} & \multirow{2}{*}{3,184} & Point-wise & HR, \(\mathrm{RelMAE}_{rel}\), \(\mathrm{RelMAE}_{abs}\) & D5 \\
 &  &  & Point-wise & \(\mathrm{RelMAE}_{ori}\) & N/A \\
\end{longtable}
}
\endgroup

\begingroup
\tiny
\captionsetup[table]{font=normalsize}
\setlength{\LTcapwidth}{\textwidth}
\setlength{\tabcolsep}{3.5pt}
\renewcommand{\arraystretch}{1.04}
{\fontsize{8.0pt}{9pt}\selectfont
\begin{longtable}{@{}llllll@{}}
\caption{Metric assignments and variant IDs for country-level Fixed Income categories.}
\label{tab:appendix_metric_assignment_country_fixed_income}\\
\toprule
Detail Category & Frequency & Eval. Points & Task & Metrics & \((p,h)\) Variant \\
\midrule
\endfirsthead
\toprule
Detail Category & Frequency & Eval. Points & Task & Metrics & \((p,h)\) Variant \\
\midrule
\endhead
\midrule
\multicolumn{6}{r}{\emph{Continued on next page}} \\
\endfoot
\bottomrule
\endlastfoot
\multirow{3}{*}{Treasury Yield} & \multirow{3}{*}{Business-daily} & \multirow{3}{*}{17,402} & Point-wise & HR, \(\mathrm{RelMAE}_{abs}\) & D5 \\
 &  &  & Point-wise & \(\mathrm{RelMAE}_{ori}\) & N/A \\
 &  &  & Cross-sectional & IC & D5 \\
\midrule
\multirow{2}{*}{Credit Spread} & \multirow{2}{*}{Business-daily} & \multirow{2}{*}{7,749} & Point-wise & HR, \(\mathrm{RelMAE}_{abs}\) & D5 \\
 &  &  & Point-wise & \(\mathrm{RelMAE}_{ori}\) & N/A \\
\midrule
\multirow{2}{*}{Bond Total Return Index} & \multirow{2}{*}{Business-daily} & \multirow{2}{*}{7,750} & Point-wise & HR, \(\mathrm{RelMAE}_{rel}\), \(\mathrm{RelMAE}_{abs}\) & D5 \\
 &  &  & Point-wise & \(\mathrm{RelMAE}_{ori}\) & N/A \\
\midrule
\multirow{2}{*}{Money Market} & \multirow{2}{*}{Business-daily} & \multirow{2}{*}{12,317} & Point-wise & HR, \(\mathrm{RelMAE}_{abs}\) & D5 \\
\end{longtable}
}
\endgroup

\begingroup
\tiny
\captionsetup[table]{font=normalsize}
\setlength{\LTcapwidth}{\textwidth}
\setlength{\tabcolsep}{7.2pt}
\renewcommand{\arraystretch}{1.15}
{\fontsize{8.0pt}{9pt}\selectfont
\begin{longtable}{@{}llllll@{}}
\caption{Metric assignments and variant IDs for country-level Market Indicators categories.}
\label{tab:appendix_metric_assignment_country_market_indicators}\\
\toprule
Detail Category & Frequency & Eval. Points & Task & Metrics & \((p,h)\) Variant \\
\midrule
\endfirsthead
\toprule
Detail Category & Frequency & Eval. Points & Task & Metrics & \((p,h)\) Variant \\
\midrule
\endhead
\midrule
\multicolumn{6}{r}{\emph{Continued on next page}} \\
\endfoot
\bottomrule
\endlastfoot
\multirow{2}{*}{Implied Volatility} & \multirow{2}{*}{Business-daily} & \multirow{2}{*}{12,879} & Point-wise & HR, \(\mathrm{RelMAE}_{rel}\), \(\mathrm{RelMAE}_{abs}\) & D5 \\
 &  &  & Point-wise & \(\mathrm{RelMAE}_{ori}\) & N/A \\
\midrule
\multirow{2}{*}{US Equity Index} & \multirow{2}{*}{Business-daily} & \multirow{2}{*}{14,310} & Point-wise & HR, \(\mathrm{RelMAE}_{rel}\), \(\mathrm{RelMAE}_{abs}\) & D5 \\
 &  &  & Point-wise & \(\mathrm{RelMAE}_{ori}\) & N/A \\
\end{longtable}
}
\endgroup

\begingroup
\captionsetup[table]{font=normalsize}
\setlength{\LTcapwidth}{\textwidth}
\setlength{\tabcolsep}{1.7pt}
\renewcommand{\arraystretch}{1.18}
{\fontsize{8.0pt}{9pt}\selectfont
\begin{longtable}{@{}llllll@{}}
\caption{Metric assignments and variant IDs for country-level Macro Indicators categories.}
\label{tab:appendix_metric_assignment_country_macro_indicators}\\
\toprule
Detail Category & Frequency & Eval. Points & Task & Metrics & \((p,h)\) Variant \\
\midrule
\endfirsthead
\toprule
Detail Category & Frequency & Eval. Points & Task & Metrics & \((p,h)\) Variant \\
\midrule
\endhead
\midrule
\multicolumn{6}{r}{\emph{Continued on next page}} \\
\endfoot
\bottomrule
\endlastfoot
\multirow{3}{*}{Inflation} & \multirow{3}{*}{Monthly} & \multirow{3}{*}{673} & Point-wise & HR, \(\mathrm{RelMAE}_{rel}\), \(\mathrm{RelMAE}_{abs}\) & M10 \\
 &  &  & Point-wise & \(\Delta\)HR, \(\Delta^2\)HR & M4 \\
 &  &  & Point-wise & \(\mathrm{RelMAE}_{ori}\) & N/A \\
\midrule
\multirow{3}{*}{Labor} & \multirow{3}{*}{Monthly} & \multirow{3}{*}{1,041} & Point-wise & HR, \(\mathrm{RelMAE}_{rel}\), \(\mathrm{RelMAE}_{abs}\) & M10 \\
 &  &  & Point-wise & \(\Delta\)HR, \(\Delta^2\)HR & M4 \\
 &  &  & Point-wise & \(\mathrm{RelMAE}_{ori}\) & N/A \\
\midrule
\multirow{3}{*}{GDP} & \multirow{3}{*}{Quarterly} & \multirow{3}{*}{50} & Point-wise & HR, \(\mathrm{RelMAE}_{rel}\), \(\mathrm{RelMAE}_{abs}\) & Q6 \\
 &  &  & Point-wise & \(\Delta\)HR, \(\Delta^2\)HR & Q3 \\
\end{longtable}
}
\endgroup

\begingroup
\tiny
\captionsetup[table]{font=normalsize}
\setlength{\LTcapwidth}{\textwidth}
\setlength{\tabcolsep}{5pt}
\renewcommand{\arraystretch}{1.15}
{\fontsize{8.0pt}{9pt}\selectfont
\begin{longtable}{@{}llllll@{}}
\caption{Metric assignments and variant IDs for individual-series ETF categories.}
\label{tab:appendix_metric_assignment_individual_etf}\\
\toprule
Detail Category & Frequency & Eval. Points & Task & Metrics & \((p,h)\) Variant \\
\midrule
\endfirsthead
\toprule
Detail Category & Frequency & Eval. Points & Task & Metrics & \((p,h)\) Variant \\
\midrule
\endhead
\midrule
\multicolumn{6}{r}{\emph{Continued on next page}} \\
\endfoot
\bottomrule
\endlastfoot
\multirow{4}{*}{ETF OHLCV Close} & \multirow{4}{*}{Business-daily} & \multirow{4}{*}{4,497,246} & Point-wise & HR, \(\mathrm{RelMAE}_{rel}\), \(\mathrm{RelMAE}_{abs}\) & D5 \\
 &  &  & Point-wise & \(\mathrm{RelMAE}_{ori}\) & N/A \\
 &  &  & Cross-sectional & IC & D5 \\
 &  &  & Portfolio & \mbox{return, vol, Sharpe, MDD} & D5 \\
\end{longtable}
}
\endgroup

\begingroup
\tiny
\captionsetup[table]{font=normalsize}
\setlength{\LTcapwidth}{\textwidth}
\setlength{\tabcolsep}{4.5pt}
\renewcommand{\arraystretch}{1.15}
{\fontsize{8.0pt}{9pt}\selectfont
\begin{longtable}{@{}llllll@{}}
\caption{Metric assignments and variant IDs for individual-series Equity categories.}
\label{tab:appendix_metric_assignment_individual_equity}\\
\toprule
Detail Category & Frequency & Eval. Points & Task & Metrics & \((p,h)\) Variant \\
\midrule
\endfirsthead
\toprule
Detail Category & Frequency & Eval. Points & Task & Metrics & \((p,h)\) Variant \\
\midrule
\endhead
\midrule
\multicolumn{6}{r}{\emph{Continued on next page}} \\
\endfoot
\bottomrule
\endlastfoot
\multirow{4}{*}{Equity OHLCV Close} & \multirow{4}{*}{Business-daily} & \multirow{4}{*}{7,963,766} & Point-wise & HR, \(\mathrm{RelMAE}_{rel}\), \(\mathrm{RelMAE}_{abs}\) & D5 \\
 &  &  & Point-wise & \(\mathrm{RelMAE}_{ori}\) & N/A \\
 &  &  & Cross-sectional & IC & D5 \\
 &  &  & Portfolio & \mbox{return, vol, Sharpe, MDD} & D5 \\
\end{longtable}
}
\endgroup

\begingroup
\tiny
\captionsetup[table]{font=normalsize}
\setlength{\LTcapwidth}{\textwidth}
\setlength{\tabcolsep}{4.5pt}
\renewcommand{\arraystretch}{1.19}
{\fontsize{8.0pt}{9pt}\selectfont
\begin{longtable}{@{}llllll@{}}
\caption{Metric assignments and variant IDs for individual-series Fundamentals categories.}
\label{tab:appendix_metric_assignment_individual_fundamentals}\\
\toprule
Detail Category & Frequency & Eval. Points & Task & Metrics & \((p,h)\) Variant \\
\midrule
\endfirsthead
\toprule
Detail Category & Frequency & Eval. Points & Task & Metrics & \((p,h)\) Variant \\
\midrule
\endhead
\midrule
\multicolumn{6}{r}{\emph{Continued on next page}} \\
\endfoot
\bottomrule
\endlastfoot
\multirow{3}{*}{Capital Expenditures} & \multirow{3}{*}{Quarterly} & \multirow{3}{*}{48,017} & Point-wise & HR, \(\mathrm{RelMAE}_{rel}\), \(\mathrm{RelMAE}_{abs}\) & Q6 \\
 &  &  & Point-wise & \(\Delta\)HR, \(\Delta^2\)HR & Q3 \\
 &  &  & Point-wise & \(\mathrm{RelMAE}_{ori}\) & N/A \\
\midrule
\multirow{3}{*}{Cash and Cash Equivalents} & \multirow{3}{*}{Quarterly} & \multirow{3}{*}{81,457} & Point-wise & HR, \(\mathrm{RelMAE}_{rel}\), \(\mathrm{RelMAE}_{abs}\) & Q6 \\
 &  &  & Point-wise & \(\Delta\)HR, \(\Delta^2\)HR & Q3 \\
 &  &  & Point-wise & \(\mathrm{RelMAE}_{ori}\) & N/A \\
\midrule
\multirow{4}{*}{EPS Basic} & \multirow{4}{*}{Quarterly} & \multirow{4}{*}{62,002} & Point-wise & HR, \(\mathrm{RelMAE}_{rel}\), \(\mathrm{RelMAE}_{abs}\) & Q6 \\
 &  &  & Point-wise & \(\Delta\)HR, \(\Delta^2\)HR & Q3 \\
\end{longtable}
}
\endgroup

\FloatBarrier

\begin{figure}[t]
\centering
\includegraphics[width=\linewidth]{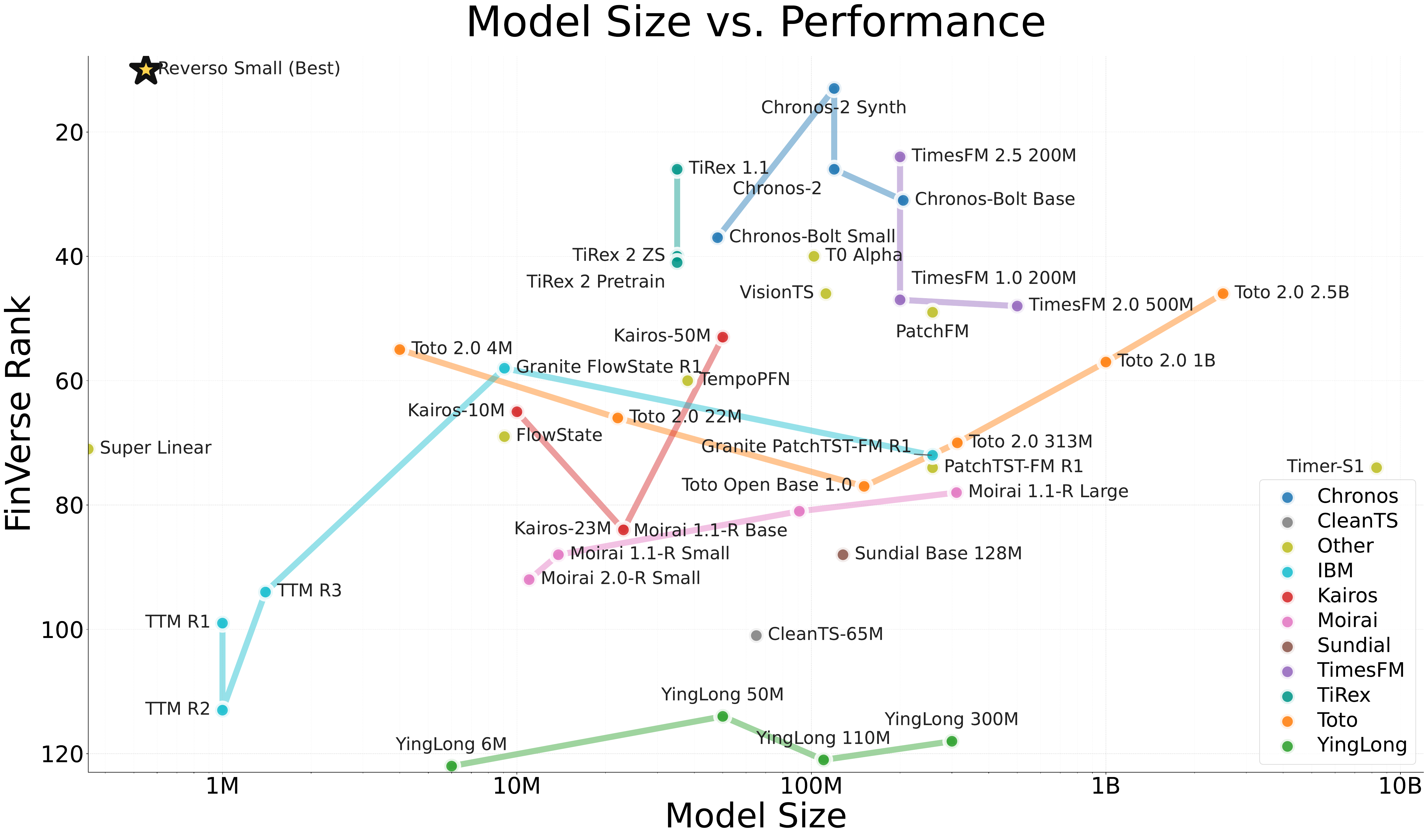}
\caption{Relationship between model size and overall FinVerse rank. Lower rank indicates better performance.}
\label{fig:model_size_performance}
\end{figure}

\begin{figure}[t]
\centering
\includegraphics[width=\linewidth]{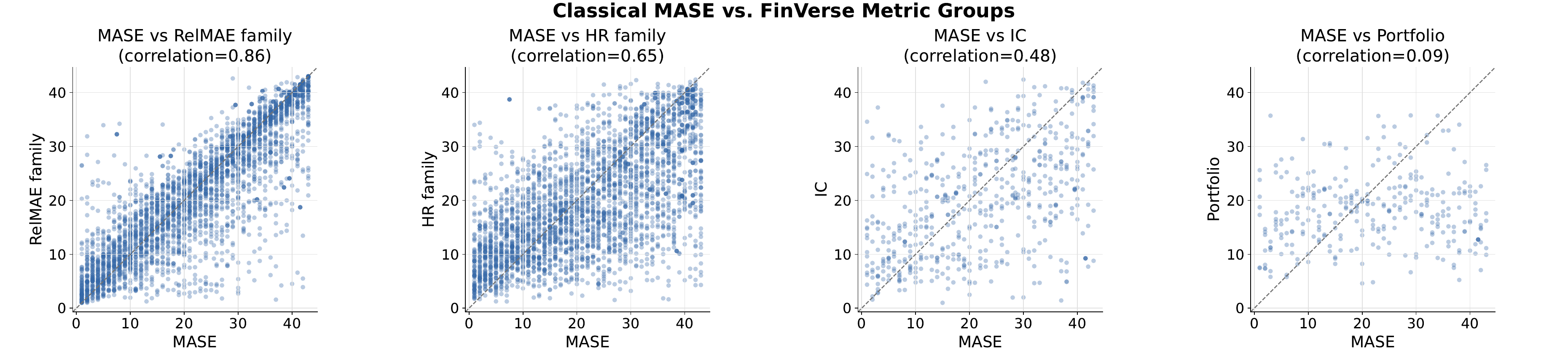}
\caption{Rank under the classical $MASE$ versus metric group under FinVerse.}
\label{fig:extra_mase_gift_rank}
\end{figure}

% \begin{figure}[t]
% \centering
% \includegraphics[width=\linewidth]{figures/extra_mase_gift_vs_finverse_rank.pdf}
% \caption{Model ranking under the assigned FinVerse metrics compared with a ranking computed from \(\mathrm{MASE}\) alone, using the same rank aggregation. The star marks the best model, which is Reverso Small under both rankings.}
% \label{fig:extra_mase_gift_rank}
% \end{figure}
%zohren: This is actually quite interesting observation, which might deserve more attention, that MASE is still a good proxy for downstream performance. 
% Wonbin: Agreed it deserves a sentence, though with a caveat. The r=0.86 in the old figure (now commented out above) was the aggregate ranking, and the new breakdown by aspect shows where it comes from: MASE tracks the RelMAE family closely (r=0.86) but the agreement drops to 0.65 for the hit ratios, 0.48 for IC and 0.09 for the portfolio metrics. So MASE is a good proxy for other point-wise error metrics and for the overall rank because point-wise carries a lot of weight in it, but a poor proxy for the decision-oriented parts. We'll say exactly that in the main text, it's a cleaner version of the paper's point than the aggregate correlation alone.

\section{Additional Results}

\subsection{Model Size and Performance}
Figure~\ref{fig:model_size_performance} examines whether overall FinVerse performance can be explained by model scale. Reverso Small achieves the best overall rank, followed by Chronos-2 Synth, TimesFM 2.5 200M, Chronos-2, and TiRex 1.1. Larger checkpoints do not consistently dominate, with several Toto, Timer, and YingLong variants ranking behind smaller or medium-sized models. These results suggest that financial decision-oriented performance cannot be explained by parameter count alone.

\subsection{Disentangling the Effects of Evaluation Data and Metrics}
\label{app:mase}

Figure~\ref{fig:gifteval_finverse_rank} compares model rankings between GIFT-Eval and FinVerse, where both the target time series and the evaluation metrics differ. To disentangle these two sources of variation, Figure~\ref{fig:extra_mase_gift_rank} instead evaluates models on the same FinVerse target series using both the FinVerse evaluation metrics and \(\mathrm{MASE}\), thereby isolating the effect of the evaluation criteria. For every model and every detail category--frequency unit, Figure~\ref{fig:extra_mase_gift_rank} plots the rank under \(\mathrm{MASE}\) against the rank under each assigned metric group. For each metric group, models are first ranked separately for each metric within a detail category--frequency unit, and these metric-wise ranks are then aggregated using the geometric mean to obtain the group-level rank. \(\mathrm{MASE}\) is nearly interchangeable with the RelMAE family (Pearson \(r=0.86\); Spearman \(\rho=0.86\); Kendall \(\tau=0.70\)), which is expected since both divide the same absolute errors by a naive-forecast scale and differ only in the choice of naive forecast. Agreement drops to \(r=0.65\) (\(\rho=0.64\), \(\tau=0.47\)) for the hit-ratio family, which evaluates the same forecasts in terms of direction rather than magnitude, to \(r=0.48\) (\(\rho=0.48\), \(\tau=0.34\)) for cross-sectional IC, and to \(r=0.09\) (\(\rho=0.08\), \(\tau=0.06\)) for the realized portfolio metrics, where the relationship all but disappears. A scaled point-forecast error is therefore a good proxy for another scaled point-forecast error but a poor proxy for decision-oriented evaluations. The ordering across the four panels also mirrors the aspect-level relationships in Figure~\ref{fig:metric_relationship_scatter}, where the point-wise and portfolio views show the weakest relationship.

\subsection{Alternative Naive Baselines}
\label{app:alt_relmae}

In this section, we evaluate five additional baselines alongside the persistence baseline used by \(\mathrm{RelMAE}_{ori}\). Let \(L\) denote the frequency-specific lookback window of Table~\ref{tab:freq_horizon_lookback}. The seasonal baselines below use a calendar seasonal period \(m'\), namely \(m'=7\) for daily, \(5\) for business-daily, \(52\) for weekly, \(12\) for monthly, and \(4\) for quarterly data, with no seasonal cycle defined for annual data. 

\begin{itemize}
    \item \textbf{Persistence} (\(\mathrm{RelMAE}_{ori}\)): repeats the last observation, \(b_{o,h}=y_{o}\). It assumes no change over the horizon.
    \item \textbf{Seasonal naive} (\(\mathrm{RelMAE}_{snaive}\)): repeats the most recent observation at the same seasonal position, \(b_{o,h}=y_{o+h-m'\lceil h/m' \rceil}\). It assumes the last seasonal cycle repeats, and is not defined for annual series.
    \item \textbf{Seasonal mean} (\(\mathrm{RelMAE}_{smean}\)): averages the observations at the same seasonal position over the last three years. It smooths the seasonal pattern instead of copying a single cycle, and like the seasonal naive it is not defined for annual series.
    \item \textbf{Random walk with drift} (\(\mathrm{RelMAE}_{drift}\)): extrapolates the slope of the lookback window, \(b_{o,h}=y_{o}+h\,(y_{o}-y_{o-L+1})/(L-1)\). It assumes the recent trend continues, i.e., a momentum view.
    \item \textbf{Lookback mean} (\(\mathrm{RelMAE}_{mean}\)): predicts the mean of the last \(L\) observations for every horizon. It assumes immediate reversion to the recent average level.
    \item \textbf{AR(1)} (\(\mathrm{RelMAE}_{ar1}\)): fits \(y_t=c+\phi y_{t-1}\) by ordinary least squares on the last three years, or on the lookback window when that is longer, and iterates it forward from \(y_o\), with \(\phi\) clipped to \([-1,1]\). It assumes gradual mean reversion at an estimated speed.
\end{itemize}

\begin{center}
\captionof{table}{Raw-value scaled error against alternative naive baselines.  Avg. is the mean of the raw metric values, and $K$ is the geometric mean of the
model ranks for that metric.}
\label{tab:extra_mase_baselines}
\vspace{0.25em}
\begingroup
\scriptsize
\setlength{\tabcolsep}{2pt}
\renewcommand{\arraystretch}{1.04}
\resizebox{\linewidth}{!}{
\begin{tabular}{@{}l|rr|rr|rr|rr|rr|rr@{}}
\toprule
\multirow{2}{*}{Model} & \multicolumn{2}{c|}{$\mathrm{RelMAE}_{ori}$} & \multicolumn{2}{c|}{$\mathrm{RelMAE}_{snaive}$} & \multicolumn{2}{c|}{$\mathrm{RelMAE}_{smean}$} & \multicolumn{2}{c|}{$\mathrm{RelMAE}_{drift}$} & \multicolumn{2}{c|}{$\mathrm{RelMAE}_{mean}$} & \multicolumn{2}{c}{$\mathrm{RelMAE}_{ar1}$} \\
\cmidrule(lr){2-3}\cmidrule(lr){4-5}\cmidrule(lr){6-7}\cmidrule(lr){8-9}\cmidrule(lr){10-11}\cmidrule(lr){12-13}
 & Avg. & $K$ & Avg. & $K$ & Avg. & $K$ & Avg. & $K$ & Avg. & $K$ & Avg. & $K$ \\
\midrule
Reverso Small & \textbf{0.981} & \textbf{5.53} & \textbf{0.807} & \textbf{4.40} & \textbf{0.609} & 4.71 & 0.805 & \textbf{5.12} & \textbf{0.770} & \textbf{5.20} & \textbf{0.932} & \textbf{5.18} \\
Chronos-2 Synth & 1.020 & 7.28 & 0.836 & 5.96 & 0.628 & 6.03 & 0.821 & 6.99 & 0.804 & 6.77 & 0.961 & 7.36 \\
TimesFM 2.5 200M & 1.002 & 8.26 & 0.820 & 6.36 & 0.620 & 6.27 & 0.826 & 7.86 & 0.780 & 7.39 & 0.953 & 7.99 \\
Chronos-2 & 1.012 & 7.69 & 0.836 & 7.65 & 0.627 & 7.57 & 0.828 & 7.41 & 0.787 & 8.44 & 0.958 & 8.02 \\
TiRex 1.1 & 0.992 & 6.35 & 0.823 & 4.85 & 0.617 & \textbf{4.68} & \textbf{0.804} & 6.30 & 0.778 & 5.81 & 0.937 & 6.03 \\
Chronos-Bolt Base & 1.049 & 11.77 & 0.863 & 11.71 & 0.653 & 11.88 & 0.863 & 11.69 & 0.815 & 12.18 & 0.998 & 12.29 \\
Chronos-Bolt Small & 1.063 & 11.38 & 0.865 & 11.19 & 0.654 & 11.22 & 0.874 & 11.85 & 0.824 & 12.00 & 1.005 & 12.42 \\
T0 Alpha & 1.207 & 19.17 & 0.962 & 18.10 & 0.713 & 17.81 & 1.009 & 19.13 & 0.907 & 18.31 & 1.135 & 19.71 \\
TiRex 2 ZS & 1.007 & 7.69 & 0.827 & 7.31 & 0.619 & 7.01 & 0.828 & 8.26 & 0.783 & 7.70 & 0.951 & 7.97 \\
TiRex 2 Pretrain & 1.001 & 7.03 & 0.824 & 6.23 & 0.616 & 5.88 & 0.823 & 7.15 & 0.781 & 6.85 & 0.946 & 6.99 \\
Toto 2.0 2.5B & 1.139 & 13.06 & 0.994 & 20.97 & 0.789 & 21.79 & 0.904 & 11.58 & 0.898 & 13.40 & 1.049 & 11.94 \\
VisionTS & 1.066 & 11.76 & 0.877 & 12.72 & 0.659 & 12.65 & 0.879 & 12.95 & 0.830 & 12.40 & 1.015 & 12.87 \\
TimesFM 1.0 200M & 1.159 & 19.79 & 0.917 & 18.10 & 0.686 & 17.41 & 0.964 & 19.36 & 0.875 & 18.22 & 1.088 & 19.73 \\
TimesFM 2.0 500M & 1.341 & 24.97 & 1.057 & 24.52 & 0.821 & 25.05 & 1.100 & 24.37 & 1.019 & 24.73 & 1.276 & 25.88 \\
PatchFM & 1.022 & 8.47 & 0.858 & 8.25 & 0.655 & 8.46 & 0.852 & 9.49 & 0.819 & 9.46 & 0.969 & 8.26 \\
Kairos-50M & 1.109 & 14.15 & 0.921 & 13.16 & 0.710 & 13.69 & 0.877 & 12.72 & 0.876 & 14.75 & 1.031 & 14.33 \\
Toto 2.0 4M & 1.084 & 13.61 & 0.919 & 15.02 & 0.704 & 15.50 & 0.877 & 13.28 & 0.853 & 13.79 & 1.007 & 12.17 \\
Toto 2.0 1B & 1.140 & 12.44 & 0.983 & 19.65 & 0.781 & 20.56 & 0.900 & 11.87 & 0.896 & 12.93 & 1.043 & 12.08 \\
Granite FlowState R1 & 1.240 & 22.53 & 1.024 & 21.77 & 0.802 & 21.74 & 1.006 & 22.21 & 0.989 & 23.12 & 1.131 & 22.00 \\
TempoPFN & 1.066 & 11.16 & 0.863 & 9.86 & 0.661 & 9.97 & 0.871 & 10.97 & 0.839 & 11.32 & 1.016 & 11.21 \\
Kairos-10M & 1.320 & 23.67 & 1.054 & 22.10 & 0.840 & 22.29 & 1.046 & 22.52 & 1.048 & 23.80 & 1.235 & 23.15 \\
Toto 2.0 22M & 1.131 & 15.02 & 0.961 & 18.67 & 0.749 & 19.38 & 0.908 & 14.19 & 0.891 & 15.18 & 1.057 & 14.26 \\
FlowState & 1.207 & 19.33 & 0.998 & 18.92 & 0.780 & 18.76 & 0.982 & 19.78 & 0.969 & 19.78 & 1.088 & 19.09 \\
Toto 2.0 313M & 1.120 & 13.37 & 0.974 & 20.22 & 0.772 & 20.89 & 0.889 & 12.38 & 0.884 & 13.98 & 1.036 & 12.10 \\
Super Linear & 1.116 & 15.79 & 0.890 & 14.41 & 0.668 & 14.16 & 0.917 & 15.34 & 0.843 & 14.53 & 1.050 & 15.89 \\
Granite PatchTST-FM R1 & 1.473 & 32.35 & 1.115 & 30.18 & 0.844 & 29.73 & 1.214 & 32.84 & 1.071 & 30.96 & 1.381 & 32.95 \\
PatchTST-FM R1 & 1.334 & 27.33 & 1.040 & 25.16 & 0.776 & 24.99 & 1.081 & 27.60 & 1.001 & 26.53 & 1.248 & 27.51 \\
Timer-S1 & 1.166 & 18.69 & 0.919 & 16.71 & 0.685 & 16.35 & 0.959 & 19.06 & 0.879 & 17.23 & 1.088 & 18.84 \\
Toto Open Base 1.0 & 1.117 & 15.22 & 0.933 & 17.83 & 0.723 & 17.69 & 0.944 & 17.01 & 0.870 & 16.87 & 1.047 & 14.37 \\
Moirai 1.1-R Large & 1.406 & 25.51 & 1.096 & 25.82 & 0.819 & 25.72 & 1.175 & 26.51 & 1.047 & 26.39 & 1.334 & 26.06 \\
Moirai 1.1-R Base & 1.348 & 26.27 & 1.056 & 27.94 & 0.799 & 27.99 & 1.124 & 28.43 & 1.023 & 28.57 & 1.277 & 27.47 \\
Kairos-23M & 1.272 & 23.72 & 1.042 & 23.64 & 0.794 & 23.94 & 1.000 & 22.45 & 1.001 & 24.18 & 1.183 & 23.52 \\
Moirai 1.1-R Small & 1.275 & 23.47 & 1.003 & 24.62 & 0.749 & 24.57 & 1.059 & 25.14 & 0.962 & 24.89 & 1.199 & 25.54 \\
Sundial Base 128M & 1.200 & 21.63 & 0.943 & 19.19 & 0.703 & 18.86 & 0.991 & 21.82 & 0.899 & 19.38 & 1.120 & 21.63 \\
Moirai 2.0-R Small & 1.369 & 25.08 & 1.106 & 27.51 & 0.808 & 27.36 & 1.136 & 27.39 & 1.027 & 26.35 & 1.275 & 25.67 \\
TTM R3 & 1.375 & 26.56 & 1.056 & 24.86 & 0.775 & 23.88 & 1.139 & 27.72 & 1.012 & 25.13 & 1.284 & 26.37 \\
TTM R1 & 1.658 & 23.63 & 1.321 & 29.49 & 0.972 & 29.70 & 1.374 & 24.13 & 1.154 & 24.73 & 1.587 & 24.54 \\
CleanTS-65M & 1.415 & 30.46 & 1.081 & 28.63 & 0.797 & 28.20 & 1.168 & 30.96 & 1.045 & 29.80 & 1.308 & 30.81 \\
TTM R2 & 1.358 & 21.14 & 1.087 & 22.99 & 0.829 & 23.11 & 1.086 & 20.48 & 0.984 & 21.29 & 1.269 & 21.73 \\
YingLong 50M & 1.782 & 39.51 & 1.312 & 38.21 & 0.970 & 37.91 & 1.469 & 39.72 & 1.283 & 38.91 & 1.655 & 39.75 \\
YingLong 300M & 1.779 & 39.18 & 1.309 & 37.86 & 0.968 & 37.56 & 1.467 & 39.38 & 1.281 & 38.58 & 1.652 & 39.41 \\
YingLong 110M & 1.774 & 38.74 & 1.303 & 37.39 & 0.964 & 37.05 & 1.462 & 38.94 & 1.278 & 38.09 & 1.647 & 38.93 \\
YingLong 6M & 1.772 & 38.64 & 1.304 & 37.31 & 0.966 & 37.10 & 1.462 & 38.83 & 1.278 & 38.10 & 1.648 & 38.86 \\
\bottomrule
\end{tabular}
}
\endgroup
\end{center}

Table~\ref{tab:extra_mase_baselines} shows that model rankings, as measured by \(K\), remain broadly stable across different naive baselines. Models that rank highly under \(\mathrm{RelMAE}_{ori}\) generally maintain strong rankings when alternative baselines are used, while lower-ranked models similarly tend to remain near the bottom. This consistency suggests that the relative comparison of forecasting models is robust to the specific choice of baseline used to scale the raw-value forecasting error. In terms of absolute performance, among the 43 evaluated models, 2 outperform the persistence baseline, 25 outperform the seasonal naive baseline, 24 outperform the random-walk-with-drift baseline, 28 outperform the lookback-mean baseline, 9 outperform the AR(1) baseline, and all 43 outperform the seasonal-mean baseline.

% Table~\ref{tab:extra_mase_baselines} shows that model rankings, as measured by $K$, remain broadly stable across different naive baselines. Models that rank highly under $\mathrm{RelMAE}_{ori}$ generally maintain strong rankings when alternative baselines are used, while lower-ranked models similarly tend to remain near the bottom. This consistency suggests that the relative comparison of forecasting models is robust to the specific choice of baseline used to scale the raw-value forecasting error.

% Hao Ni: one more thing, how does the benchmarking results change over different time periods. whether the model comparison results are consistent over time.
% Wonbin: We had the same question, this table and the subsection right below are our attempt at it. Rankings are recomputed for each origin year 2020-2025 and compared with each other and with the pooled ranking. Agreement between years is moderate (pairwise Spearman 0.42 to 0.80, mean 0.61), so the ordering is reasonably robust but not fixed. 2020 is the noisiest year as you'd expect, and a few models jump around quite a bit. We'll point to this from the main text and say something about the year-to-year variation.

\clearpage

\begin{center}
\captionof{table}{Overall FinVerse rank sum by forecast-origin year.}
\label{tab:yearly_rank_sum}
\vspace{0.25em}
\begingroup
\scriptsize
\setlength{\tabcolsep}{10pt}
\renewcommand{\arraystretch}{1.04}
\resizebox{\linewidth}{!}{
\begin{tabular}{@{}l|r|rrrrrr@{}}
\toprule
Model & Full & 2020 & 2021 & 2022 & 2023 & 2024 & 2025 \\
\midrule
Reverso Small & \textbf{10} & 28 & 16 & 20 & \textbf{11} & \textbf{12} & \textbf{8} \\
Chronos-2 Synth & 13 & 39 & 23 & \textbf{14} & 13 & 19 & 14 \\
TimesFM 2.5 200M & 24 & 33 & 52 & 30 & 20 & 49 & 28 \\
Chronos-2 & 26 & 45 & \textbf{11} & 32 & 39 & 40 & 32 \\
TiRex 1.1 & 26 & 31 & 40 & \textbf{14} & 27 & 28 & 33 \\
Chronos-Bolt Base & 31 & 23 & 78 & 32 & 53 & 59 & 66 \\
Chronos-Bolt Small & 37 & \textbf{22} & 54 & 45 & 51 & 73 & 69 \\
T0 Alpha & 40 & 53 & 46 & 32 & 44 & 53 & 21 \\
TiRex 2 ZS & 40 & 47 & 54 & 39 & 40 & 48 & 34 \\
TiRex 2 Pretrain & 41 & 41 & 55 & 35 & 48 & 46 & 26 \\
Toto 2.0 2.5B & 46 & 75 & 73 & 87 & 49 & 40 & 50 \\
VisionTS & 46 & 32 & 52 & 62 & 54 & 66 & 43 \\
TimesFM 1.0 200M & 47 & 50 & 27 & 44 & 64 & 65 & 77 \\
TimesFM 2.0 500M & 48 & 63 & 90 & 38 & 39 & 40 & 44 \\
PatchFM & 49 & 41 & 61 & 59 & 47 & 49 & 39 \\
Kairos-50M & 53 & 49 & 78 & 47 & 74 & 66 & 64 \\
Toto 2.0 4M & 55 & 84 & 54 & 84 & 55 & 35 & 81 \\
Toto 2.0 1B & 57 & 67 & 62 & 83 & 71 & 34 & 34 \\
Granite FlowState R1 & 58 & 104 & 90 & 73 & 41 & 60 & 86 \\
TempoPFN & 60 & 91 & 24 & 93 & 73 & 44 & 49 \\
Kairos-10M & 65 & 75 & 73 & 85 & 69 & 60 & 59 \\
Toto 2.0 22M & 66 & 90 & 57 & 77 & 63 & 60 & 63 \\
FlowState & 69 & 108 & 74 & 87 & 53 & 56 & 87 \\
Toto 2.0 313M & 70 & 68 & 68 & 99 & 73 & 61 & 44 \\
Super Linear & 71 & 38 & 56 & 71 & 50 & 68 & 36 \\
Granite PatchTST-FM R1 & 72 & 39 & 49 & 71 & 110 & 88 & 83 \\
PatchTST-FM R1 & 74 & 40 & 54 & 80 & 111 & 87 & 90 \\
Timer-S1 & 74 & 60 & 75 & 72 & 63 & 80 & 71 \\
Toto Open Base 1.0 & 77 & 83 & 105 & 96 & 94 & 76 & 51 \\
Moirai 1.1-R Large & 78 & 90 & 56 & 63 & 85 & 69 & 57 \\
Moirai 1.1-R Base & 81 & 82 & 64 & 54 & 72 & 84 & 73 \\
Kairos-23M & 84 & 81 & 77 & 99 & 101 & 67 & 78 \\
Moirai 1.1-R Small & 88 & 95 & 90 & 60 & 61 & 78 & 67 \\
Sundial Base 128M & 88 & 57 & 67 & 66 & 74 & 82 & 89 \\
Moirai 2.0-R Small & 92 & 100 & 94 & 68 & 74 & 109 & 88 \\
TTM R3 & 94 & 63 & 86 & 72 & 48 & 75 & 99 \\
TTM R1 & 99 & 85 & 108 & 70 & 91 & 104 & 115 \\
CleanTS-65M & 101 & 69 & 81 & 85 & 74 & 80 & 116 \\
TTM R2 & 113 & 69 & 105 & 116 & 83 & 82 & 113 \\
YingLong 50M & 114 & 111 & 90 & 97 & 120 & 103 & 112 \\
YingLong 300M & 118 & 105 & 98 & 101 & 124 & 122 & 114 \\
YingLong 110M & 121 & 101 & 80 & 84 & 124 & 116 & 128 \\
YingLong 6M & 122 & 111 & 91 & 102 & 108 & 105 & 107 \\
\bottomrule
\end{tabular}
}
\endgroup
\end{center}

\subsection{Stability of the Model Comparison over Time}
\label{app:stability}

The results reported so far pool all forecast origins from 2020 onward. Because financial regimes change, a natural question is whether the model comparison itself is stable over time. We therefore repeat the evaluation separately for the forecast origins of each calendar year, using the same metric assignment and rank aggregation. We then compare the rankings across individual years as well as with the full-period ranking. The pairwise correlations between yearly rankings indicate moderate agreement across evaluation years, with Spearman correlations ranging from \(0.42\) to \(0.80\) (mean \(\rho=0.61\)). These results suggest that, while model rankings exhibit some year-to-year variation, the relative performance of models in FinVerse is reasonably robust across evaluation periods rather than being driven by a single year.

\subsection{Detailed Model Rankings}
\label{app:detailed_rankings}

Table~\ref{tab:candidate_overall} presents the overall FinVerse leaderboard, including each model’s overall rank, aspect-level ranks, and corresponding \(K\) values. The overall rank is computed as the sum of the point-wise, IC, and portfolio placement ranks, while \(K\) denotes the geometric mean of metric-level ranks within each evaluation aspect. Tables~\ref{tab:candidate_scope_rank} and~\ref{tab:candidate_frequency_rank} further break down model performance by financial data scope (Country, Individual, and Inter-country) and sampling frequency, respectively. Finally, Tables~\ref{tab:detail_category_model_performance}–\ref{tab:detail_category_model_performance_9} provide a finer-grained breakdown by detail category, where each value represents the geometric mean of the corresponding metric-level model ranks and lower values indicate better performance. The detail-category results are split across multiple tables for readability. Repeated values across the four YingLong checkpoints are expected because their forecasts are almost constant over the horizon for the lower-frequency series regardless of checkpoint size.
% \subsection{Stability of the Model Comparison over Time}

% Table~\ref{tab:candidate_overall} reports the overall leaderboard used in the main analysis. It shows each model's overall FinVerse rank, computed as the sum of the point-wise, IC, and portfolio placement ranks, together with the aspect-level ranks and the corresponding K values. The K values are geometric means of metric-level ranks within each evaluation aspect. Table~\ref{tab:candidate_scope_rank} reports scope-wise performance. It aggregates metric-level ranks separately within the Country, Individual, and Inter-country scopes, showing whether a model performs consistently across different financial data scopes or is relatively stronger in a particular scope. 
% %
% Table~\ref{tab:candidate_frequency_rank} reports frequency-wise performance. It aggregates metric-level ranks separately across different sampling frequencies, showing whether a model performs consistently across frequencies or is relatively stronger at particular frequencies.
% %
% Tables~\ref{tab:detail_category_model_performance}--\ref{tab:detail_category_model_performance_9} report model performance at the detail-category level. Each value is the geometric mean of the metric-level model ranks for the corresponding detail category; lower values are better. For readability, the results are divided into grouped tables.

\clearpage

\begin{center}
\captionof{table}{Overall and aspect-wise model rankings. Overall is the sum of the point-wise, IC, and portfolio placement ranks. K columns report the geometric mean of metric ranks within each aspect.}
\label{tab:candidate_overall}
\vspace{0.25em}
\begingroup
\scriptsize
\setlength{\tabcolsep}{6pt}
\renewcommand{\arraystretch}{1.06}
\resizebox{\linewidth}{!}{
\begin{tabular}{@{}r l r rr rr rr@{}}
\toprule
\multirow{2}{*}{Rank} & \multirow{2}{*}{Model} & \multirow{2}{*}{Overall Rank} & \multicolumn{2}{c}{Point} & \multicolumn{2}{c}{IC} & \multicolumn{2}{c}{Portfolio} \\
\cmidrule(lr){4-5}\cmidrule(lr){6-7}\cmidrule(lr){8-9}
 &  &  & Rank & K & Rank & K & Rank & K \\
\midrule
1 & Reverso Small & \textbf{10} & 6 & 9.39 & \textbf{1} & \textbf{7.95} & 3 & 11.42 \\
2 & Chronos-2 Synth & 13 & 9 & 11.38 & 2 & 8.25 & 2 & 10.58 \\
3 & TimesFM 2.5 200M & 24 & 7 & 9.91 & 4 & 10.09 & 13 & 15.47 \\
4 & Chronos-2 & 26 & 5 & 9.18 & 13 & 15.32 & 8 & 14.55 \\
5 & TiRex 1.1 & 26 & \textbf{1} & \textbf{6.73} & 3 & 9.32 & 22 & 16.94 \\
6 & Chronos-Bolt Base & 31 & 11 & 12.03 & 15 & 15.72 & 5 & 13.52 \\
\bottomrule
\end{tabular}
}
\endgroup
\end{center}

\clearpage
\begin{center}
\captionof{table}{Scope-wise rank aggregation. Lower values are better.}
\label{tab:candidate_scope_rank}
\vspace{0.25em}
\begingroup
\scriptsize
\setlength{\tabcolsep}{23pt}
\renewcommand{\arraystretch}{1.04}
\resizebox{\linewidth}{!}{
\begin{tabular}{@{}l|rrr@{}}
\toprule
Model & Country & Individual & Inter-country \\
\midrule
Reverso Small & 9.32 & 9.91 & 9.45 \\
Chronos-2 Synth & 12.33 & 11.35 & \textbf{7.06} \\
TimesFM 2.5 200M & 9.88 & 10.71 & 11.15 \\
Chronos-2 & 9.14 & 9.77 & 12.01 \\
TiRex 1.1 & \textbf{6.04} & 8.92 & 12.44 \\
Chronos-Bolt Base & 12.11 & 11.95 & 13.28 \\
Chronos-Bolt Small & 12.11 & 12.03 & 13.22 \\
T0 Alpha & 11.41 & 15.38 & 19.60 \\
TiRex 2 ZS & 7.44 & 9.88 & 13.99 \\
TiRex 2 Pretrain & 6.55 & 10.66 & 12.49 \\
Toto 2.0 2.5B & 14.82 & 16.46 & 8.92 \\
VisionTS & 11.15 & \textbf{8.57} & 19.16 \\
TimesFM 1.0 200M & 16.46 & 14.09 & 19.51 \\
TimesFM 2.0 500M & 30.31 & 14.47 & 20.79 \\
PatchFM & 7.63 & 18.20 & 13.02 \\
Kairos-50M & 14.93 & 20.04 & 14.71 \\
Toto 2.0 4M & 13.03 & 19.89 & 12.21 \\
Toto 2.0 1B & 14.08 & 16.74 & 8.97 \\
Granite FlowState R1 & 15.02 & 22.79 & 15.33 \\
TempoPFN & 14.78 & 18.07 & 8.29 \\
Kairos-10M & 19.75 & 26.64 & 15.41 \\
Toto 2.0 22M & 15.07 & 18.30 & 11.56 \\
FlowState & 15.40 & 25.96 & 15.77 \\
Toto 2.0 313M & 14.34 & 16.53 & 10.24 \\
Super Linear & 18.55 & 11.79 & 20.76 \\
Granite PatchTST-FM R1 & 29.88 & 20.43 & 25.04 \\
PatchTST-FM R1 & 26.23 & 20.10 & 24.94 \\
Timer-S1 & 18.67 & 14.45 & 21.31 \\
Toto Open Base 1.0 & 19.52 & 16.95 & 16.99 \\
Moirai 1.1-R Large & 27.11 & 21.46 & 25.34 \\
Moirai 1.1-R Base & 26.48 & 20.82 & 25.70 \\
Kairos-23M & 20.02 & 23.56 & 21.17 \\
Moirai 1.1-R Small & 24.85 & 20.82 & 24.67 \\
Sundial Base 128M & 19.31 & 17.10 & 22.94 \\
Moirai 2.0-R Small & 22.08 & 22.15 & 24.22 \\
TTM R3 & 28.26 & 19.02 & 24.24 \\
TTM R1 & 23.62 & 18.21 & 24.85 \\
CleanTS-65M & 26.39 & 23.77 & 26.39 \\
TTM R2 & 21.89 & 17.80 & 23.43 \\
YingLong 50M & 36.28 & 27.45 & 26.27 \\
YingLong 300M & 36.45 & 27.38 & 27.02 \\
YingLong 110M & 36.29 & 28.00 & 27.07 \\
YingLong 6M & 35.97 & 27.47 & 27.23 \\
\bottomrule
\end{tabular}
}
\endgroup
\end{center}

\clearpage
\begin{center}
\captionof{table}{Frequency-wise rank aggregation. Lower values are better.}
\label{tab:candidate_frequency_rank}
\vspace{0.25em}
\begingroup
\scriptsize
\setlength{\tabcolsep}{7pt}
\renewcommand{\arraystretch}{1.04}
\resizebox{\linewidth}{!}{
\begin{tabular}{@{}l|rrrrrr@{}}
\toprule
Model & Business-daily & Daily & Weekly & Monthly & Quarterly & Annual \\
\midrule
Reverso Small & 8.08 & 11.50 & 12.08 & 9.92 & 9.43 & 13.37 \\
Chronos-2 Synth & 7.38 & \textbf{4.34} & 8.08 & 11.57 & 15.41 & 18.90 \\
TimesFM 2.5 200M & 10.84 & 12.80 & 9.23 & 9.36 & 11.25 & 8.80 \\
Chronos-2 & 13.05 & 11.55 & 10.17 & 8.73 & 8.83 & 14.46 \\
TiRex 1.1 & 11.51 & 14.31 & 8.54 & \textbf{4.66} & 10.03 & 12.99 \\
Chronos-Bolt Base & 13.32 & 11.41 & 8.00 & 13.06 & 10.66 & 9.01 \\
Chronos-Bolt Small & 12.41 & 15.09 & 9.77 & 12.80 & 11.17 & 13.95 \\
\bottomrule
\end{tabular}
}
\endgroup
\end{center}

\clearpage
\begin{center}
\captionof{table}{Detail-category-wise rank aggregation (1/9).}
\label{tab:detail_category_model_performance}
\vspace{0.25em}
\scriptsize
\setlength{\tabcolsep}{1pt}
\renewcommand{\arraystretch}{1.04}
\resizebox{\linewidth}{!}{
\begin{tabular}{@{}l|cccccccc@{}}
\toprule
Model & Activity & Banking & \makecell{Bond Total\\Return Index} & Breakeven & \makecell{Capacity Utilization\\by Industry} & \makecell{Capital\\Expenditures} & \makecell{Cash and Cash\\Equivalents} & \makecell{Commodity\\ETF Broad} \\
\midrule
Reverso Small & 8.12 & 17.04 & 6.55 & 9.45 & 5.84 & 29.05 & 14.48 & 10.94 \\
Chronos-2 Synth & 12.13 & 10.29 & 7.25 & 8.90 & 15.75 & 14.10 & 9.58 & 5.29 \\
TimesFM 2.5 200M & 6.17 & 11.52 & 15.76 & 7.81 & 7.93 & 19.28 & 10.17 & 10.05 \\
Chronos-2 & 8.57 & 5.46 & 17.46 & 6.96 & 8.73 & 17.78 & 12.50 & 7.38 \\
TiRex 1.1 & 3.54 & 7.02 & 11.99 & 6.94 & 3.82 & 10.59 & 3.76 & 12.11 \\
Chronos-Bolt Base & 11.05 & 12.18 & 7.08 & 13.03 & 9.28 & 14.89 & 11.85 & 18.08 \\
Chronos-Bolt Small & 14.99 & 18.28 & \textbf{2.09} & 18.32 & 11.79 & 19.50 & 12.36 & 13.64 \\
\bottomrule
\end{tabular}
}
\end{center}

\clearpage
\begin{center}
\captionof{table}{Detail-category-wise rank aggregation (2/9).}
\label{tab:detail_category_model_performance_2}
\vspace{0.25em}
\scriptsize
\setlength{\tabcolsep}{2pt}
\renewcommand{\arraystretch}{1.04}
\resizebox{\linewidth}{!}{
\begin{tabular}{@{}l|cccccccc@{}}
\toprule
Model & \makecell{Commodity\\ETF Single} & \makecell{Commodity\\ETF Thematic} & \makecell{Corporate\\Yield} & \makecell{Country\\Index} & Credit & \makecell{Credit\\Spread} & \makecell{Crypto Pair\\Close Top 100} & \makecell{Current\\Account} \\
\midrule
Reverso Small & 5.90 & \textbf{3.62} & 12.13 & \textbf{5.56} & 10.75 & 9.39 & 11.50 & 8.29 \\
Chronos-2 Synth & 5.15 & 11.37 & 16.02 & 14.81 & 9.27 & 13.86 & \textbf{4.34} & 19.30 \\
TimesFM 2.5 200M & 12.31 & 8.37 & 11.73 & 15.24 & 15.44 & 6.28 & 12.80 & 5.96 \\
Chronos-2 & 15.20 & 23.81 & 14.18 & 19.82 & 8.11 & 24.15 & 11.55 & 14.31 \\
TiRex 1.1 & 7.64 & 7.58 & 12.40 & 18.06 & \textbf{2.29} & 9.19 & 14.31 & 11.32 \\
Chronos-Bolt Base & 15.93 & 7.23 & 13.64 & 13.80 & 15.65 & 15.24 & 11.41 & 10.16 \\
Chronos-Bolt Small & 16.29 & 4.17 & 15.28 & 10.26 & 16.27 & 18.23 & 15.09 & 9.92 \\
\bottomrule
\end{tabular}
}
\end{center}

\clearpage
\begin{center}
\captionof{table}{Detail-category-wise rank aggregation (3/9).}
\label{tab:detail_category_model_performance_3}
\vspace{0.25em}
\scriptsize
\setlength{\tabcolsep}{4pt}
\renewcommand{\arraystretch}{1.04}
\resizebox{\linewidth}{!}{
\begin{tabular}{@{}l|cccccccc@{}}
\toprule
Model & \makecell{EPS\\Basic} & \makecell{EPS\\Diluted} & \makecell{ETF OHLCV\\Close} & \makecell{Energy\\Activity} & \makecell{Equity\\OHLCV Close} & External & \makecell{FDI\\Flows} & \makecell{FX\\Pair} \\
\midrule
Reverso Small & 10.25 & 9.94 & 9.71 & 10.45 & 6.12 & 5.24 & 16.02 & 17.50 \\
Chronos-2 Synth & 14.68 & 16.21 & 6.34 & 13.11 & \textbf{4.07} & 13.92 & 8.03 & 8.08 \\
TimesFM 2.5 200M & 10.56 & 10.55 & 11.66 & 13.78 & 11.25 & 9.99 & 14.61 & 7.43 \\
Chronos-2 & 16.44 & 16.58 & 10.11 & 4.57 & 13.46 & 7.65 & 20.22 & 13.75 \\
TiRex 1.1 & 10.74 & 11.58 & 11.95 & \textbf{2.61} & 12.08 & \textbf{3.30} & 10.31 & 12.30 \\
Chronos-Bolt Base & 17.81 & 18.17 & 10.39 & 12.02 & 16.41 & 13.81 & 13.78 & 10.26 \\
Chronos-Bolt Small & 21.08 & 21.18 & 10.91 & 6.91 & 18.39 & 14.92 & 20.38 & 10.70 \\
\bottomrule
\end{tabular}
}
\end{center}

\clearpage
\begin{center}
\captionof{table}{Detail-category-wise rank aggregation (4/9).}
\label{tab:detail_category_model_performance_4}
\vspace{0.25em}
\scriptsize
\setlength{\tabcolsep}{6pt}
\renewcommand{\arraystretch}{1.04}
\resizebox{\linewidth}{!}{
\begin{tabular}{@{}l|cccccccc@{}}
\toprule
Model & \makecell{Financial\\Flows} & Fiscal & \makecell{Forward\\Inflation} & Freight & \makecell{Future\\Front} & GDP & Housing & \makecell{Housing\\Price} \\
\midrule
Reverso Small & 6.71 & 12.38 & 12.19 & 10.05 & 12.26 & 6.68 & 17.03 & 17.75 \\
Chronos-2 Synth & 20.61 & 15.24 & 11.47 & 7.69 & \textbf{4.85} & 24.92 & 8.44 & 11.96 \\
TimesFM 2.5 200M & 13.22 & \textbf{7.94} & 14.40 & 8.29 & 17.09 & 11.99 & \textbf{1.91} & 9.89 \\
Chronos-2 & 4.56 & 10.78 & 13.60 & 7.93 & 10.78 & 12.41 & 9.74 & 8.70 \\
TiRex 1.1 & 18.05 & 15.74 & 15.63 & 9.18 & 13.11 & 11.56 & 10.60 & 3.81 \\
Chronos-Bolt Base & 5.73 & 12.81 & 21.20 & 12.20 & 17.27 & 17.41 & 9.32 & 20.46 \\
Chronos-Bolt Small & \textbf{3.92} & 14.79 & 27.94 & 9.26 & 21.14 & 19.32 & 4.08 & 21.57 \\
\bottomrule
\end{tabular}
}
\end{center}

\clearpage
\begin{center}
\captionof{table}{Detail-category-wise rank aggregation (5/9).}
\label{tab:detail_category_model_performance_5}
\vspace{0.25em}
\scriptsize
\setlength{\tabcolsep}{2pt}
\renewcommand{\arraystretch}{1.04}
\resizebox{\linewidth}{!}{
\begin{tabular}{@{}l|cccccccc@{}}
\toprule
Model & \makecell{Implied\\Volatility} & Income & \makecell{Industrial Production\\by Industry} & Inflation & \makecell{Inflation\\Expectation} & \makecell{International\\Investment Position} & Inventory & \makecell{Inventory\\Stock} \\
\midrule
Reverso Small & 8.83 & 10.17 & 5.72 & 12.01 & 17.02 & 12.45 & 7.42 & 14.02 \\
Chronos-2 Synth & 25.94 & 12.09 & 13.42 & 10.30 & 11.50 & 24.92 & 11.26 & 7.21 \\
TimesFM 2.5 200M & 9.94 & 14.78 & 9.38 & 9.37 & 12.05 & 20.60 & 6.33 & 5.04 \\
Chronos-2 & 23.36 & 8.77 & 10.28 & 4.31 & 13.76 & 7.25 & 11.09 & 10.00 \\
TiRex 1.1 & 9.17 & \textbf{2.44} & 2.92 & \textbf{2.27} & 7.60 & 19.84 & 2.27 & 9.66 \\
Chronos-Bolt Base & 28.22 & 12.78 & 11.28 & 12.31 & 15.01 & 7.95 & 12.96 & 8.91 \\
Chronos-Bolt Small & 27.23 & 13.97 & 12.19 & 16.66 & 11.34 & 8.59 & 16.69 & 10.07 \\
\bottomrule
\end{tabular}
}
\end{center}

\clearpage
\begin{center}
\captionof{table}{Detail-category-wise rank aggregation (6/9).}
\label{tab:detail_category_model_performance_6}
\vspace{0.25em}
\scriptsize
\setlength{\tabcolsep}{4pt}
\renewcommand{\arraystretch}{1.04}
\resizebox{\linewidth}{!}{
\begin{tabular}{@{}l|cccccccc@{}}
\toprule
Model & Labor & \makecell{Labor by\\Industry} & \makecell{Long\\Term Debt} & Money & \makecell{Money\\Market} & \makecell{Mortgage\\Rate} & \makecell{Net\\Income} & \makecell{Operating\\Cash Flow} \\
\midrule
Reverso Small & 7.33 & 5.77 & 10.37 & 10.68 & 3.39 & 9.72 & 7.16 & 23.15 \\
Chronos-2 Synth & 10.12 & 10.69 & 17.91 & 11.47 & 12.29 & 9.55 & 7.61 & 13.45 \\
TimesFM 2.5 200M & 8.01 & 13.42 & 12.42 & 15.84 & 4.14 & 22.24 & 6.25 & 13.54 \\
Chronos-2 & 7.69 & 7.21 & \textbf{3.00} & 8.23 & 15.08 & 10.43 & 19.76 & 21.14 \\
TiRex 1.1 & \textbf{2.35} & \textbf{2.28} & 9.84 & \textbf{2.68} & 7.58 & 7.15 & \textbf{4.77} & \textbf{8.55} \\
Chronos-Bolt Base & 15.94 & 15.05 & 5.13 & 10.97 & 12.80 & 6.83 & 17.61 & 22.32 \\
Chronos-Bolt Small & 10.62 & 13.37 & 4.05 & 10.48 & 15.03 & 9.35 & 20.70 & 23.20 \\
\bottomrule
\end{tabular}
}
\end{center}

\clearpage
\begin{center}
\captionof{table}{Detail-category-wise rank aggregation (7/9).}
\label{tab:detail_category_model_performance_7}
\vspace{0.25em}
\scriptsize
\setlength{\tabcolsep}{2pt}
\renewcommand{\arraystretch}{1.04}
\resizebox{\linewidth}{!}{
\begin{tabular}{@{}l|cccccccc@{}}
\toprule
Model & \makecell{Operating\\Income} & \makecell{Portfolio\\Flows} & Productivity & \makecell{Regional\\Fed} & \makecell{Regional\\Index} & Revenue & Saving & Sentiment \\
\midrule
Reverso Small & 7.28 & 6.84 & 9.14 & 11.71 & 9.01 & 7.42 & 6.10 & 11.21 \\
Chronos-2 Synth & 9.63 & \textbf{5.22} & 24.00 & \textbf{4.58} & 3.78 & 13.23 & 19.92 & 13.94 \\
TimesFM 2.5 200M & 5.76 & 7.43 & 16.52 & 7.72 & 5.36 & 8.01 & 6.31 & 14.45 \\
Chronos-2 & 18.82 & 11.00 & 5.20 & 10.60 & 4.06 & 9.77 & 11.74 & 8.59 \\
TiRex 1.1 & \textbf{5.52} & 6.50 & 19.29 & 7.87 & 18.06 & 8.04 & 11.22 & 3.97 \\
Chronos-Bolt Base & 15.81 & 10.79 & 6.69 & 14.33 & 12.39 & 11.04 & 14.59 & 9.15 \\
Chronos-Bolt Small & 18.79 & 9.40 & 8.34 & 20.34 & 11.21 & 12.65 & 10.04 & 8.32 \\
\bottomrule
\end{tabular}
}
\end{center}

\clearpage
\begin{center}
\captionof{table}{Detail-category-wise rank aggregation (8/9).}
\label{tab:detail_category_model_performance_8}
\vspace{0.25em}
\scriptsize
\setlength{\tabcolsep}{4pt}
\renewcommand{\arraystretch}{1.04}
\resizebox{\linewidth}{!}{
\begin{tabular}{@{}l|cccccccc@{}}
\toprule
Model & Spending & Spot & \makecell{Stockholders\\Equity} & Stress & \makecell{TIC\\Flows} & \makecell{TIPS\\Yield} & \makecell{Term\\Spread} & \makecell{Total\\Assets} \\
\midrule
Reverso Small & 6.85 & 8.13 & 10.06 & 15.52 & 11.26 & \textbf{2.32} & 7.89 & 8.06 \\
Chronos-2 Synth & 13.55 & 8.82 & 24.03 & 16.01 & 25.04 & 10.16 & 4.67 & 19.17 \\
TimesFM 2.5 200M & 8.02 & 13.63 & 14.21 & 7.89 & 17.32 & 19.73 & 16.29 & 12.89 \\
Chronos-2 & 12.08 & 10.47 & \textbf{2.14} & 7.09 & 4.44 & 19.78 & 22.41 & \textbf{2.34} \\
TiRex 1.1 & 3.97 & 10.46 & 11.87 & 10.58 & 18.91 & 8.45 & 15.17 & 10.86 \\
Chronos-Bolt Base & 18.33 & 17.60 & 6.40 & 12.43 & 4.81 & 10.07 & 7.84 & 5.43 \\
Chronos-Bolt Small & 23.24 & 18.34 & 3.52 & \textbf{7.04} & 7.46 & 14.48 & 5.84 & 3.58 \\
\bottomrule
\end{tabular}
}
\end{center}

\clearpage
\begin{center}
\captionof{table}{Detail-category-wise rank aggregation (9/9).}
\label{tab:detail_category_model_performance_9}
\vspace{0.25em}
% \
\setlength{\tabcolsep}{4pt}
\renewcommand{\arraystretch}{1.04}
\resizebox{0.55\linewidth}{!}{
\begin{tabular}{@{}l|ccc@{}}
\toprule
Model & \makecell{Total\\Liabilities} & \makecell{Treasury\\Yield} & \makecell{US Equity\\Index} \\
\midrule
Reverso Small & 7.85 & \textbf{2.46} & 14.37 \\
Chronos-2 Synth & 17.40 & 18.00 & 6.99 \\
TimesFM 2.5 200M & 13.39 & 18.24 & 11.04 \\
Chronos-2 & 3.42 & 29.36 & 5.45 \\
TiRex 1.1 & 10.17 & 7.04 & 12.47 \\
Chronos-Bolt Base & 6.46 & 9.46 & 9.22 \\
Chronos-Bolt Small & 5.07 & 11.05 & 8.03 \\
T0 Alpha & 15.08 & 12.56 & 29.15 \\
TiRex 2 ZS & 7.39 & 9.01 & 21.04 \\
TiRex 2 Pretrain & 10.79 & 9.60 & 18.63 \\
Toto 2.0 2.5B & 21.19 & 6.38 & \textbf{1.71} \\
VisionTS & \textbf{2.49} & 20.89 & 20.87 \\
TimesFM 1.0 200M & 15.92 & 26.89 & 22.26 \\
TimesFM 2.0 500M & 13.77 & 20.04 & 34.38 \\
PatchFM & 31.19 & 5.89 & 11.27 \\
Kairos-50M & 17.83 & 15.23 & 8.76 \\
Toto 2.0 4M & 24.79 & 16.87 & 10.45 \\
Toto 2.0 1B & 25.08 & 7.82 & 4.34 \\
Granite FlowState R1 & 22.90 & 5.90 & 19.43 \\
TempoPFN & 16.16 & 9.25 & 2.31 \\
Kairos-10M & 20.50 & 11.28 & 16.06 \\
Toto 2.0 22M & 22.09 & 17.63 & 8.71 \\
FlowState & 28.32 & 14.96 & 19.77 \\
Toto 2.0 313M & 20.52 & 12.52 & 4.34 \\
Super Linear & 9.83 & 32.41 & 27.53 \\
Granite PatchTST-FM R1 & 27.97 & 39.56 & 34.81 \\
PatchTST-FM R1 & 30.68 & 36.51 & 32.25 \\
Timer-S1 & 18.25 & 22.63 & 27.27 \\
Toto Open Base 1.0 & 13.74 & 14.81 & 16.67 \\
Moirai 1.1-R Large & 22.96 & 25.75 & 38.05 \\
Moirai 1.1-R Base & 24.61 & 16.24 & 40.18 \\
Kairos-23M & 19.02 & 27.70 & 24.18 \\
Moirai 1.1-R Small & 20.04 & 22.00 & 36.01 \\
Sundial Base 128M & 21.72 & 31.36 & 30.57 \\
Moirai 2.0-R Small & 16.55 & 10.20 & 33.09 \\
TTM R3 & 26.17 & 22.75 & 21.84 \\
TTM R1 & 14.60 & 29.44 & 30.81 \\
CleanTS-65M & 33.01 & 40.32 & 36.96 \\
TTM R2 & 14.68 & 25.74 & 22.65 \\
YingLong 50M & 37.83 & 37.41 & 34.25 \\
YingLong 300M & 37.83 & 39.04 & 34.24 \\
YingLong 110M & 37.83 & 37.97 & 34.61 \\
YingLong 6M & 37.83 & 37.55 & 33.24 \\
\bottomrule
\end{tabular}
}
\end{center}

\end{document}